\documentclass[runningheads]{llncs}

\usepackage{eccv}

\usepackage{eccvabbrv}

\usepackage{graphicx}
\usepackage{booktabs}

\usepackage[accsupp]{axessibility}  

\usepackage{xcolor}         
\usepackage{graphicx}
\usepackage{epstopdf}
\usepackage{multirow}
\usepackage{colortbl}
\usepackage[pagebackref=true,breaklinks=true,letterpaper=true,colorlinks,bookmarks=false]{hyperref}
\usepackage{subcaption}
\usepackage{amsmath} 
\usepackage{amssymb}            
\usepackage{mathrsfs}            
\usepackage{bm}
\usepackage{wrapfig}
\usepackage{bbm}

\usepackage{hyperref}

\usepackage{orcidlink}
\usepackage{algorithm}
\usepackage{algorithmic}
\usepackage{marvosym}
\usepackage{inconsolata}
\usepackage{ulem}
\makeatletter
\newcommand*\bigcdot{\mathpalette\bigcdot@{.5}}
\newcommand*\bigcdot@[2]{\mathbin{\vcenter{\hbox{\scalebox{#2}{$\m@th#1\bullet$}}}}}
\makeatother

\definecolor{cvprblue}{rgb}{0.21,0.49,0.74}
\usepackage{hyperref}
\definecolor{c2}{HTML}{FBD9BD}
\definecolor{c3}{HTML}{fe793d}
\definecolor{c4}{HTML}{eedeb0}
\definecolor{pp}{HTML}{BC7FCD}
\definecolor{bb}{HTML}{CDE8E5}
\definecolor{rouse}{rgb}{0.981,0.961,0.941}

\begin{document}

\title{UnfoldLDM: Degradation-Aware Unfolding \\
with Iterative Latent Diffusion Priors \\
for Blind Image Restoration}

\titlerunning{Abbreviated paper title}

\author{Chunming He$^{1,*}$\,,
        Rihan Zhang$^{1,*}$\thanks{Equal Contribution, $\dagger$ Corresponding Author}\,~,
        Zheng Chen$^{2}$\,,\\
        Bowen Yang$^{3}$\,, 
	{Chengyu Fang}$^{4}$\,, 
    {Yunlong Lin}$^{5}$ \,, \\ 
    Yulun Zhang$^{2}$\,,
            {Fengyang Xiao}$^{1,\dagger}$ \,, 
        and {Sina Farsiu}$^{1,\dagger}$\\
        $^1$Duke University,
	$^2$Shanghai Jiao Tong University,  
        $^3$Peking University, \\
        $^4$Tsinghua University, 
        $^5$Xiamen University. 
}

\authorrunning{He et al.}

\institute{
\email{chunming.he@duke.edu}}

\maketitle
\begin{center}
\includegraphics[width=\textwidth]{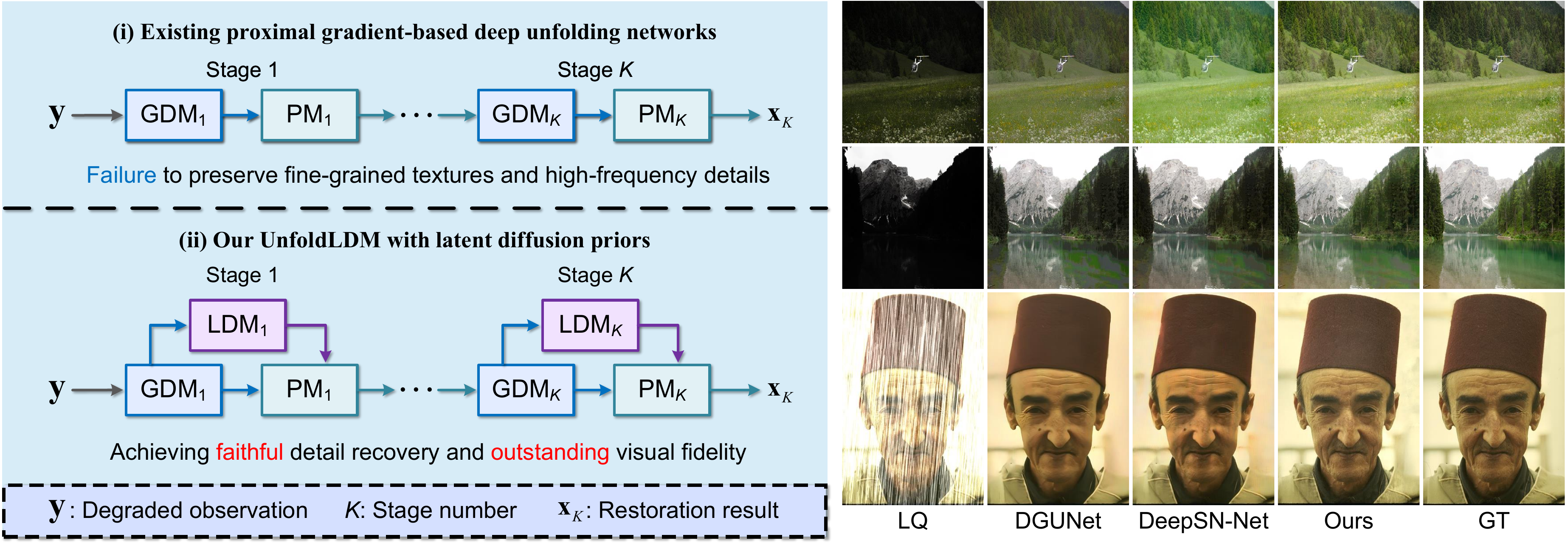}\vspace{-3mm} 
\captionof{figure}{Comparison between existing proximal gradient DUN-based methods (\textit{e.g.}, DGUNet~\cite{mou2022deep} and DeepSN-Net~\cite{deng2025deepsn}) and our UnfoldLDM. UnfoldLDM better resists unknown degradation and eliminates the over-smoothing bias of existing DUNs.} \label{Fig:ArchitectureCompare}
\end{center}

\begin{abstract}
Deep unfolding networks (DUNs) combine the interpretability of model-based methods with the learning ability of deep networks, yet remain limited for blind image restoration (BIR). Existing DUNs suffer from: 
(1) \textbf{Degradation-specific dependency}, as their optimization frameworks are tied to a known degradation model, making them unsuitable for BIR tasks; and (2) \textbf{Over-smoothing bias}, resulting from the direct feeding of gradient descent outputs, dominated by low-frequency content, into the proximal term, suppressing fine textures. 
To overcome these issues, we propose UnfoldLDM to integrate DUNs with latent diffusion model (LDM) for BIR. 
In each stage, UnfoldLDM employs a multi-granularity degradation-aware (MGDA) module as the gradient descent step. MGDA models BIR as an unknown degradation estimation problem and estimates both the holistic degradation matrix and its decomposed forms, enabling robust degradation removal. 
For the proximal step, we design a degradation-resistant LDM (DR-LDM) to extract compact degradation-invariant priors from the MGDA output. Guided by this prior, an over-smoothing correction transformer (OCFormer) explicitly recovers high-frequency components and enhances texture details. This unique combination ensures the final result is degradation-free and visually rich. 
Experiments show that our UnfoldLDM achieves a leading place on eight diverse BIR tasks and benefits downstream tasks. Moreover, our design is compatible with existing DUN-based methods, serving as a plug-and-play framework. Code will be released.
  \keywords{Blind image restoration \and Deep unfolding network \and Diffusion model}
\end{abstract}
\setlength{\abovedisplayskip}{2pt}
\setlength{\belowdisplayskip}{2pt}
\section{Introduction}
\label{sec:intro}
Blind image restoration (BIR) aims to recover high‑quality images from unknown degradations \cite{he2023hqg,xia2023diffir}.
It plays a vital role in numerous applications, including photography \cite{he2023reti}, medical imaging \cite{park2025deep}, and downstream vision tasks \cite{he2023camouflaged,deng2022pcgan}. Traditional methods based on handcrafted priors are interpretable but struggle to generalize to real‑world degradations \cite{ju2024all}, whereas learning‑based methods achieve superior performance but often lack interpretability \cite{he2023reti,chen2024binarized}.

Deep unfolding networks (DUNs) have emerged as a promising paradigm to bridge this gap~\cite{sun2016deep}. 
By unfolding the iterative optimization into a multi-stage network, DUNs inherit the model-based interpretability while leveraging the learning-based representational power. 
Among them, proximal-gradient (PG)-based DUNs are widely adopted for their flexibility and effectiveness~\cite{he2025unfoldir,he2025run}. As shown in~\cref{Fig:ArchitectureCompare}, a typical PG‑based DUN alternates between a gradient descent step derived from the observation model and a proximal operator parameterized by a learnable prior, enforcing data fidelity while enhancing perceptual quality.

However, existing PG-based DUNs face two challenges: 
\textbf{(1) Degradation-specific dependency}. Most are designed for a particular degradation type (\textit{e.g.}, deblurring or low-light enhancement) and rely on known physical priors, making them unsuitable for complex or mixed degradations. 
\textbf{(2) Over-smoothing bias}. The gradient descent step derives its updates from the data fidelity term, whose gradients are dominated by low-frequency residuals in degraded images. As a result, the intermediate estimate passed to the proximal operator carries predominantly low-frequency content, creating an information bottleneck that starves the proximal operator of the high-frequency cues needed for texture recovery. This bias accumulates across stages, producing over-smoothed results with diminished structural fidelity (see~\cref{Fig:ArchitectureCompare}).

To overcome the aforementioned problems, we propose UnfoldLDM, which first integrates DUNs with the latent diffusion model (LDM) for BIR. As illustrated in Fig.~\ref{Fig.framework}, each stage in UnfoldLDM has two components: (i) a multi-granularity degradation-aware (MGDA) module serving as the gradient descent term, and (ii) a proximal design comprising a Degradation-Resistant LDM (DR-LDM) and an over-smoothing correction transformer (OCFormer).

In MGDA, we formulate BIR as an unknown degradation estimation problem by jointly estimating the holistic degradation matrix and its decomposed factors. The consistency between these two representations is guaranteed by an intra-stage degradation-aware (ISDA) loss. Solving these two forms in tandem ensures both scalability and stability, thereby facilitating robust degradation removal.

For the proximal step, DR-LDM extracts degradation-invariant priors from the MGDA output by performing diffusion in a low-dimensional latent space, distilling high-frequency cues into a compact representation while filtering out spatially correlated artifacts that would otherwise mislead restoration.
Guided by this prior, OCFormer explicitly restores fine-grained texture details suppressed in the earlier period. As unfolding progresses, MGDA increasingly captures degradation patterns, while DR-LDM and OCFormer progressively refine texture recovery, ensuring results that are both degradation-free and visually rich.

Our contributions are summarized as follows:

\noindent(1) We propose UnfoldLDM, the first method that integrates DUNs with latent diffusion priors for BIR, alleviating degradation-specific dependence and over-smoothing bias inherent in existing DUN-based methods. 

\noindent(2) We propose the MGDA module, which jointly estimates holistic and decomposed degradation forms. An ISDA loss is further introduced to ensure consistent degradation estimation, enabling robust and stable restoration.

\noindent(3) We design a DR-LDM to extract a compact degradation-invariant prior, which guides OCFormer to explicitly recover high-frequency textures.

\noindent(4) Experiments across eight diverse BIR tasks and downstream applications validate our superiority and generalizability. Moreover, DR-LDM serves as a plug-and-play module that yields consistent improvements when integrated into existing DUN-based methods across six representative tasks.

\vspace{-1mm}
\section{Related Works}\vspace{-1mm}
\label{sec:related}

\noindent \textbf{DUN-based image restoration}. 
DUNs~\cite{he2023degradation,he2025unfoldir} translate iterative optimization into trainable networks for low-level vision tasks such as deblurring\cite{mou2022deep}, super-resolution\cite{zhang2020deep}, and low-light enhancement\cite{he2025unfoldir}.
Most DUNs adopt a proximal gradient scheme, combining a gradient descent step with a proximal operator.
However, when applied to BIR, they face two major challenges: (\textit{i}) degradation-specific designs limit generalization to unknown degradations, and (\textit{ii}) the strong coupling between descent and proximal terms biases recovery toward low-frequency components, yielding over-smoothed results.
This highlights the need for degradation-agnostic DUNs that preserve fine structural details.

\noindent \textbf{Prior-guided image restoration}. Image priors are crucial for constraining solution spaces in restoration.
Classical handcrafted priors, such as total variation \cite{chambolle2004algorithm}, lack robustness under real‑world degradations, while deep generative priors from GANs \cite{he2023hqg}, VAEs \cite{deng2022pcgan}, and diffusion models~\cite{yi2023diff,xia2023diffir} capture richer natural statistics and yield realistic results.
However, these priors can be misled by heavily degraded inputs, leading to false restoration.
A promising direction is to couple degradation‑aware modeling with generative priors, enabling texture‑preserving restoration while maintaining data fidelity constraints.
Following this direction, our UnfoldLDM integrates latent diffusion priors into a degradation‑aware DUN, effectively addressing the above issues.

\noindent \textbf{Diffusion models for image restoration.} 
Pixel-space diffusion approaches such as 
IR-SDE~\cite{luo2023image} and 
GSAD~\cite{jinhui2023global} achieve high visual quality 
but at substantial cost. Latent-space methods reduce this 
overhead: StableSR~\cite{wang2024exploiting} adapts 
pretrained Stable Diffusion with ControlNet for 
super-resolution, OSEDiff~\cite{wu2024one} achieves 
one-step blind SR via score distillation, and 
Reti-Diff~\cite{he2023reti} introduces Retinex-guided 
latent diffusion for illumination degradation. Despite 
their effectiveness, these methods lack explicit 
degradation modeling and apply priors without the 
iterative refinement inherent to optimization-based 
frameworks. Our UnfoldLDM bridges this gap by embedding 
a compact latent diffusion prior within a multi-stage 
degradation-aware DUN, where the prior is progressively 
conditioned on cleaner estimates across stages.

\section{Restoration Model and Optimization}
\label{sec:model}
\noindent \textbf{Restoration model}.
Blind image restoration (BIR) is an ill-posed problem. Given the degraded observation $\mathbf{y}$, the degradation process can be formulated 
\begin{equation}
    \mathbf{y} = \mathbf{D} \mathbf{x}  + \mathbf{n},
\end{equation}
where $\mathbf{y},\mathbf{x} \in \mathbb{R}^{c\times h \times w}$ and $\mathbf{x}$ is the latent clean image, $\mathbf{D}\in \mathbb{R}^{c\times hw \times hw}$ is the unknown degradation matrix with per-channel operators to capture channel-varying distortions, and $\mathbf{n}$ is additive noise.
The objective is to recover the optimal clean image by minimizing the following energy:
\begin{equation}
 L(\mathbf{x}) =  \; \frac{1}{2}\|\mathbf{y} - \mathbf{D} \mathbf{x}\|^2_2 + \lambda \phi(\mathbf{x}),
\end{equation}
where $\|\bigcdot\|^2_2$ is the $l_2$-norm, $\phi(\cdot)$ denotes a regularization term learned by deep networks to encode prior knowledge, and $\lambda$ is a trade-off parameter.

To address the complexity of blind settings and improve modeling efficiency, we introduce a structured decomposition of the holistic degradation matrix $\mathbf{D}$ into two spatially decoupled matrices, $\mathbf{W}  \in \mathbb{R}^{c\times h \times h}$ and $\mathbf{M}   \in \mathbb{R}^{c\times w \times w}$:
\begin{equation}
    \mathbf{D} = \mathbf{M}^T \otimes \mathbf{W},
\end{equation}
where $\otimes$ denotes the Kronecker product. 

This factorization is more efficient than directly learning $\mathbf{D}$, whose dimension grows quadratically with image resolution. Also, the decomposed ones enhance structure awareness: $\mathbf{W}$ captures spatial transformations, while $\mathbf{M}$ models spectral or directional distortions. Integrating both the holistic form $\mathbf{D}$ and the decomposed form $(\mathbf{W}, \mathbf{M})$ enables expressive modeling of complex degradations while maintaining efficiency.
The final objective function is:
\begin{equation}\label{eq:Model}
L(\mathbf{x}) = \; \frac{1}{2}\|\mathbf{y} - \mathbf{D} \mathbf{x}\|^2_2 + \frac{1}{2}\|\mathbf{y} - \mathbf{W} \mathbf{x} \mathbf{M}\|^2_2 + \lambda \phi(\mathbf{x}).
\end{equation}
\begin{figure*}[t!]
  \setlength{\abovecaptionskip}{-0.4cm}
  \centering
    \includegraphics[width=1\linewidth]{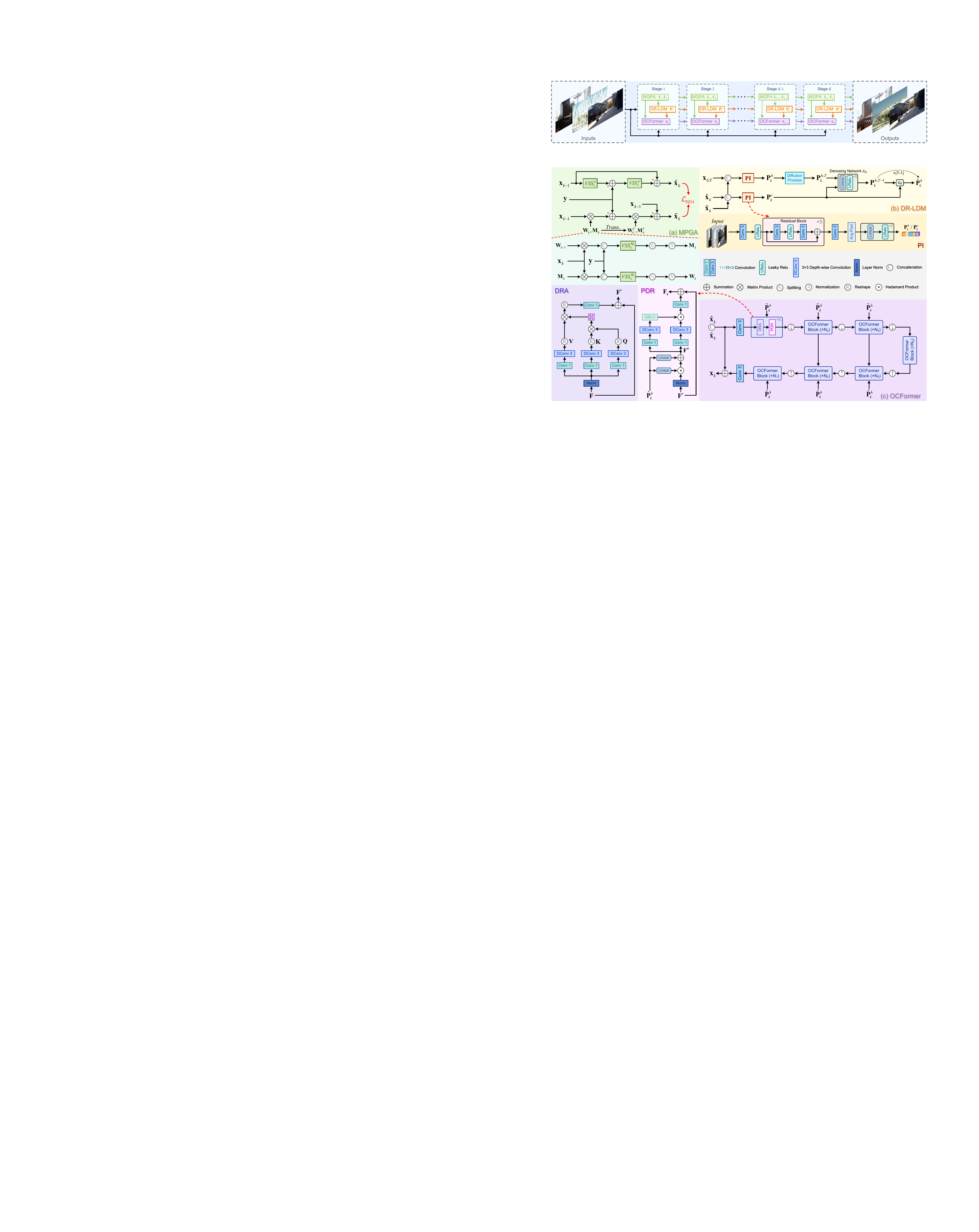}
    \newline
    \caption{Framework of our proposed UnfoldLDM.}
    \label{Fig.framework}
\vspace{-0.4cm}
\end{figure*}
\noindent \textbf{Model optimization}. To minimize~\cref{eq:Model}, we use proximal gradient algorithm~\cite{mou2022deep}. At the $k^{th}$ iteration ($1\!\leq\! k\! \leq\! K$), the optimization proceeds as 
\begin{equation}\hspace{-3mm}\label{eq:optim}
 \mathbf{x}_k \!=\! \arg \min_{\mathbf{x}} \; \frac{1}{2}\|\mathbf{y} \!-\! \mathbf{D} \mathbf{x}\|^2_2 \!+\! \frac{1}{2}\|\mathbf{y} \!-\! \mathbf{W} \mathbf{x} \mathbf{M}\|^2_2 \!+\! \lambda \phi(\mathbf{x}).
\end{equation}

Unlike conventional practices with a single gradient and proximal step, we address both holistic and decomposed degradations, requiring two gradient descent steps (one per fidelity term) and one proximal step per iteration:
\begin{equation}
 g(\mathbf{x})=\frac{1}{2}\|\mathbf{y} - \mathbf{D} \mathbf{x}\|^2_2, \ \ h(\mathbf{x}) = \frac{1}{2}\|\mathbf{y} - \mathbf{W} \mathbf{x} \mathbf{M}\|^2_2.
\end{equation}
Denoting the two intermediate gradient updates as $\hat{\mathbf{x}}_k$ and $\tilde{\mathbf{x}}_k$, we have:
\begin{equation}\label{eq:gradient1}
\begin{aligned}
     \hat{\mathbf{x}}_k &= \mathbf{x}_{k-1}- \eta\nabla g(\mathbf{x}_{k-1}),\\
     &= \mathbf{x}_{k-1}-\eta \mathbf{D}^T( \mathbf{D}\mathbf{x}_{k-1} - \mathbf{y}),
\end{aligned}
\end{equation}
\begin{equation}\label{eq:gradient2}
\begin{aligned}
 \tilde{\mathbf{x}}_k &= \mathbf{x}_{k-1}-\gamma\nabla h(\mathbf{x}_{k-1}), \\
 &=\mathbf{x}_{k-1}-\gamma \mathbf{W}^T( \mathbf{W} \mathbf{x}_{k-1} \mathbf{M} - \mathbf{y})\mathbf{M}^T,
\end{aligned}
\end{equation}
\begin{equation}\label{eq:proxim}
 {\mathbf{x}}_k = \text{prox}_{\lambda,\phi}(\hat{\mathbf{x}}_k,\tilde{\mathbf{x}}_k),
\end{equation}
where $\eta$ and $\gamma$ are two step-size parameters. This enables both global degradation modeling through $\mathbf{D}$ and structure-aware refinement via $(\mathbf{W}, \mathbf{M})$, resulting in complementary gradient updates that jointly improve restoration quality.

\begin{figure*}[t!]
  \setlength{\abovecaptionskip}{-0.4cm}
  \centering
    \includegraphics[width=\linewidth]{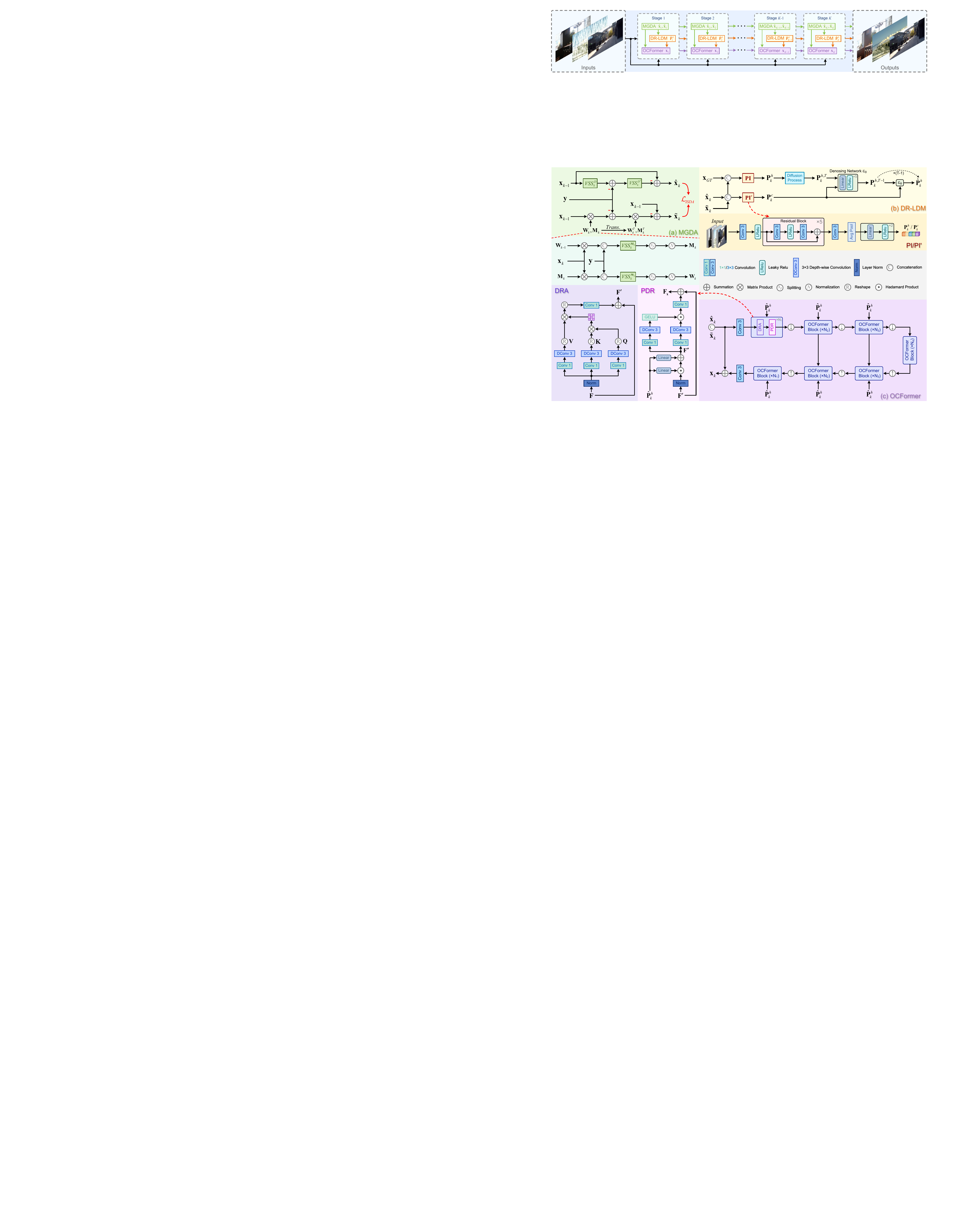}
    \newline
    \caption{Details of MGDA, DR-LDM, and OCFormer at the $k^{th}$ stage.}
    \label{Fig.framework1}
\vspace{-0.6cm}
\end{figure*}
\section{UnfoldLDM}\vspace{-2mm}
\label{sec:dun}
We unfold the iterative optimization process into a multi-stage network, UnfoldLDM.
As shown in~\cref{Fig.framework,Fig.framework1}, each stage comprises a multi-granularity degradation-aware (MGDA) module as the gradient descent term and a proximal operator composed of a degradation-resistant latent diffusion model (DR-LDM) and an over-smoothing correction transformer (OCFormer).
As the unfolding proceeds, these modules collaboratively refine degradation estimation and enhance texture recovery. 
In practice, UnfoldLDM is trained in two phases to encourage the DR-LDM to generate high-quality priors.

\vspace{-2mm}
\subsection{Phase I: Pretrain UnfoldLDM}\vspace{-1mm}
We first pretrain UnfoldLDM to encode clean images into compact priors through a Prior Inference (PI) module, and use these priors to guide OCFormer for detail reconstruction.
This allows the network to learn what type of prior information benefits fine-detail recovery. The extracted ``GT'' priors then supervise DR-LDM in producing similar high-quality priors at Phase II when the input is degraded.

\noindent \textbf{MGDA}. MGDA follows the gradient terms in \cref{eq:gradient1,eq:gradient2}, with the learnable $\eta$ and $\gamma$. Given that degradation is unknown, MGDA adopts a data-driven formulation to estimate degradation operators and gradients. At stage $k$, two Siamesed Visual State Space (VSS) blocks~\cite{guo2024mambair}, termed $VSS^{\mathbf{D}}_k(\bigcdot)$ and ${VSS^{\mathbf{D}^T}_k}(\bigcdot)$, are used to simulate $\mathbf{D}$ and $\mathbf{D}^T$. VSS is adopted because holistic degradation modeling demands global spatial reasoning.
Thus,~\cref{eq:gradient1} can be rewritten:
\begin{equation}\label{eq:x_hat}
     \hat{\mathbf{x}}_k = \mathbf{x}_{k-1}- \eta_k {VSS^{\mathbf{D}^T}_k}( VSS^{\mathbf{D}}_k(\mathbf{x}_{k-1}) - \mathbf{y}).
\end{equation}

Estimating the decomposed matrices $(\mathbf{W},\mathbf{M})$ is challenging due to their mutual dependence. To handle this, we reformulate the optimization problem:
\begin{equation}\hspace{-3mm}
 \tilde{\mathbf{x}}_k \!=\! \arg \min_{\tilde{\mathbf{x}}} \;  \frac{1}{2}\|\mathbf{y} - \mathbf{W} \tilde{\mathbf{x}} \mathbf{M}\|^2_2 \!+\! \psi_{\mathbf{W}}(\mathbf{W}) \!+\! \psi_{\mathbf{M}}(\mathbf{M}),
\end{equation}
where $\psi_{\mathbf{W}}(\bigcdot)$ and $\psi_{\mathbf{M}}(\bigcdot)$ are regularization terms for $\mathbf{W}$ and $\mathbf{M}$. 
Since $\mathbf{W}$ and $\mathbf{M}$ are physically coupled, we solve them alternatively,
effectively preventing gradient oscillations.
We replace $\psi_{\mathbf{W}}(\bigcdot)$ and $\psi_{\mathbf{M}}(\bigcdot)$ with learnable update functions $\mathcal{D}_{\mathbf{M}}(\bigcdot)$ and $\mathcal{D}_{\mathbf{W}}(\bigcdot)$, transforming hand-crafted priors into learnable ones.
By integrating their physical interdependence within the fidelity term, we get
\begin{equation}
\setlength{\abovedisplayskip}{0pt}
\setlength{\belowdisplayskip}{0pt}
    \mathbf{M}_k = \mathcal{D}_{\mathbf{M}}(\mathbf{y}, \mathbf{W}_{k-1}\mathbf{x}_{k-1}), \quad \mathbf{W}_k = \mathcal{D}_{\mathbf{W}}(\mathbf{y}, \mathbf{x}_{k-1}\mathbf{M}_k),
\end{equation} 
initialized with $\mathbf{M}_0=N(\mathbf{y}^T\mathbf{y})$ and $\mathbf{W}_0=N(\mathbf{y}\mathbf{y}^T)$, where $N(\bigcdot)$ denotes $l_2$ normalization for consistent scaling. 
To compute $\mathbf{M}_k$, we estimate two half-matrix blocks, $\mathbf{M}_k^1$ and $\mathbf{M}_k^2$, as its left and right projections via a VSS module that extracts non-local correlations from the concatenation of $\mathbf{y}$ and $\mathbf{W}_{k-1}\mathbf{x}_k$: 
\begin{equation}
    \mathbf{M}_k^1, \mathbf{M}_k^2 = S_p(VSS_k^{\mathbf{M}_k}(\text{conca}(\mathbf{y},\mathbf{W}_{k-1}\mathbf{x}_{k-1}))),
\end{equation}
where $S_p(\bigcdot)$ denotes channel-wise splitting, ensuring $\mathbf{M}_k^1, \mathbf{M}_k^2 \in \mathbb{R}^{c\times w \times w}$.
We then aggregate their correlations in the embedding space, formulated as:
\begin{equation}
    \mathbf{M}_k = N(({\mathbf{M}_k^1})^T \mathbf{M}_k^2),
\end{equation}
which encourages $\mathbf{M}_k$ to capture statistical correlations and directional dependencies between $\mathbf{y}$ and $\mathbf{W}_{k-1}\mathbf{x}_k$, forming a structured approximation of degradation. $\mathbf{W}_k$ can be calculated similarly.
Then, the decomposed update is:
\begin{equation}\label{eq:x_tilde}
 \tilde{\mathbf{x}}_k = \mathbf{x}_{k-1}-\gamma_k \mathbf{W}_k^T( \mathbf{W}_k \mathbf{x}_{k-1} \mathbf{M}_k - \mathbf{y})\mathbf{M}_k^T,
\end{equation}
with $\hat{\mathbf{x}}_k$ captures coarse global consistency (\cref{eq:x_hat}), while $\tilde{\mathbf{x}}_k$ refines local structure (\cref{eq:x_tilde}), forming a optimization balancing scalability and precision.

\noindent \textbf{OCFormer}. Given the MGDA outputs $\hat{\mathbf{x}}_k$ and $\tilde{\mathbf{x}}_k$, we concatenate them with the clean image $\mathbf{x}_{GT}$ and feed the combined one into the PI module (see~\cref{Fig.framework1} (b)), a network commonly used for learning compact latent priors~\cite{he2023hqg}:
\begin{equation}
    \mathbf{P}^h_k = \text{PI}(\text{conca}(\hat{\mathbf{x}}_k,\tilde{\mathbf{x}}_k,\mathbf{x}_{GT})),
\end{equation}
where $\mathbf{P}^h_k \in \mathbb{R}^{C_p}$ is a compact vector. By including $\mathbf{x}_{GT}$, the prior $\mathbf{P}^h_k$ captures the differences between the current estimate and the clean target, establishing a high-quality reference prior space that DR-LDM learns to approximate from degraded inputs alone in Phase II.
$\mathbf{P}^h_k$ guides OCFormer to refine $\hat{\mathbf{x}}_k$ and $\tilde{\mathbf{x}}_k$:
\begin{equation}
    \mathbf{x}_k = \text{OCFormer}(\hat{\mathbf{x}}_k,\tilde{\mathbf{x}}_k,\mathbf{P}^h_k),
\end{equation}
which, alike~\cref{eq:proxim}, serves a learnable proximal operator. OCFormer adopts a U-shaped structure with specialized blocks comprising degradation-resistant attention (DRA) and prior-guided detail recovery (PDR) modules.

We first extract features $\mathbf{F}$ from the concatenation of $\hat{\mathbf{x}}_k$ and $\tilde{\mathbf{x}}_k$ via $3\times 3$ convolutions (the stage index $k$ is omitted). DRA employs self-attention to capture complementary information between the two feature sets. Specifically, queries, keys, and values are projected via $\mathbf{W}$ combining $1\times 1$ point-wise and $3\times 3$ depth-wise convolutions, formulated as:
\begin{equation}\label{eq:qkv}
    \mathbf{Q} = \mathbf{W}_Q \mathbf{F}, \ \ \mathbf{K} = \mathbf{W}_K \mathbf{F}, \ \ \mathbf{V} = \mathbf{W}_V \mathbf{F}. 
\end{equation}
The self-attention output $\mathbf{F}'$, with Softmax $S(\bigcdot)$ and a scaling factor $I$, is:
\begin{equation}
    \mathbf{F}' = S(\mathbf{Q}\mathbf{K}^T/I)\cdot \mathbf{V}+\mathbf{F}.
\end{equation}
To recover lost details, PDR integrates the high-quality prior $\mathbf{P}^h_k$:
\begin{equation}
    \mathbf{F}_{{\mathbf{x}}}= \mathbf{F}' + \text{GELU} (\mathbf{W}_G \mathbf{F}'')\odot \mathbf{W}_H \mathbf{F}'',
\end{equation}\vspace{-2mm}
\begin{equation}
    \mathbf{F}''= Linear_1(\mathbf{P}^h_k)\odot LN(\mathbf{F}')+Linear_2(\mathbf{P}^h_k),
\end{equation}
where $\mathbf{W}_G$ and $\mathbf{W}_H$ share projection structures of~\cref{eq:qkv}. $\text{GELU}(\bigcdot)$, $Linear(\bigcdot)$, and $LN(\bigcdot)$ are GELU~\cite{he2023reti}, linear layers, and layer normalization. The prior reweights latent features, enhancing fine structures that DUNs often oversmooth.

\noindent \textbf{Optimization}. After $K$ stages,
we obtain the final result $\mathbf{x}_K$. Given $\mathbf{x}_{GT}$, the basic reconstruction loss follows the practice of~\cite{fang2024real}:
\begin{equation}
    \mathcal{L}_{Rec} = \|\mathbf{x}_K-\mathbf{x}_{GT}\|_1.
\end{equation}
Apart from $\mathcal{L}_{Rec}$, MGDA estimates both holistic and decomposed degradations, which are theoretically equivalent but empirically complementary.
To encourage mutual consistency without impairing this complementarity, we introduce the Intra-Stage Degradation-Aware (ISDA) loss:
\begin{equation}
    \mathcal{L}_{ISDA} = \sum_{k=2}^K \frac{1}{2^{K-k}}\|\hat{\mathbf{x}}_k-\tilde{\mathbf{x}}_k\|_1, 
\end{equation}
applied from the second stage onward.
The total loss is defined as:
\begin{equation}
    \mathcal{L}^I_{Total} = \mathcal{L}_{Rec} +  \mathcal{L}_{ISDA}. 
\end{equation}

\vspace{-1mm}
\subsection{Phase II: Optimize DR-LDM}\vspace{-1mm}
In phase II, DR-LDM is trained to generate high-quality priors from $(\hat{\mathbf{x}}_k, \tilde{\mathbf{x}}_k)$. The predicted prior $\hat{\mathbf{P}}^h_k$ aims to match the ${\mathbf{P}}^h_k$ extracted by the pretrained PI.

\noindent \textbf{Diffusion process}. In this process, we first use PI to extract the clean prior ${\mathbf{P}}^h_k$, which serves as the initialization of the forward Markov chain: ${\mathbf{P}}^h_k = {\mathbf{P}}^{h,0}_k$. 
Gaussian noise is then added across $T$ steps. At step $t$, the forward process is:
\begin{equation}\hspace{-3mm}
    q\left(\mathbf{P}^{h,t}_k |\mathbf{P}^{h,{t-1}}_k \right)=\mathcal{N}\left(\mathbf{P}^{h,t}_k;\sqrt{1-\beta^t}\mathbf{P}^{h,{t-1}}_k , \beta^t \mathbf{I} \right),
\end{equation}
where $t \in [1, T]$. $\mathbf{P}^{h,t}_k$ denotes the noisy prior at time step $t$ and stage $k$. $\beta^t$ is a predefined variance schedule controlling the noise strength and $\mathcal{N}$ is the Gaussian distribution. Following~\cite{kingma2013auto}, we define $\alpha^t = 1-\beta^t$ and $\bar{{\alpha}}^t=\prod_{i=1}^t\alpha^i$, which leads to the simplified marginal distribution:
\begin{equation}
  q\left(\mathbf{P}^{h,t}_k |\mathbf{P}^{h,0}_k \right)=\mathcal{N}\left(\mathbf{P}^{h,t}_k;\sqrt{\bar{\alpha}^t}\mathbf{P}^{h,0}_k , (1-\bar{\alpha}^t) \mathbf{I} \right).
\end{equation}

\noindent \textbf{Reverse process}. In this process, DR-LDM recovers the clean prior from a Gaussian noise sample. It begins with a random initialization $\mathbf{P}^{h,T}_k\sim \mathcal{N}(0,\mathbf{I})$ and progressively denoises it to $\mathbf{P}^{h,0}_k$. Given the clean prior $\mathbf{P}_k^{h,0}$ 
(available only during training), the true posterior 
of each reverse step is: 
\begin{equation}\hspace{-3mm}
    p\!\left(\mathbf{P}^{h,{t-1}}_k |\mathbf{P}^{h,t|0}_k\!\right)\!= \!\mathcal{N}\!\left(\mathbf{P}^{h,{t-1}}_k;\boldsymbol{\mu}^t(\mathbf{P}^{h,t|0}_k), (\boldsymbol{\sigma}^t)^2\mathbf{I} \!\right),
\end{equation}
where $\mathbf{P}^{h,t|0}_k = (\mathbf{P}^{h,t}_k, \mathbf{P}^{h,0}_k)$, $\boldsymbol{\mu}^t(\mathbf{P}^{h,t|0}_k)=\frac{1}{\sqrt{\alpha^t}}(\mathbf{P}^{h,t}_k - \frac{1-\alpha^t}{\sqrt{1-\bar{\alpha}^t}}\boldsymbol{\epsilon})$, and $(\boldsymbol{\sigma}^t)^2=\frac{1-\bar{\alpha}^{t-1}}{1-\bar{\alpha}^t}\beta^t$. $\boldsymbol{\epsilon}$ is Gaussian noise estimated by a denoising network $\boldsymbol{\epsilon}_\theta (\bigcdot)$ of 5 Linear+LeakyReLU layers. Since $\mathbf{P}_k^{h,0}$ depends on 
$\mathbf{x}_{GT}$ and is unavailable at inference, 
we introduce an auxiliary prior extraction module 
$\mathrm{PI}'$ to provide a conditional surrogate. $\mathrm{PI}'$ computes a conditional cue 
$\mathbf{P}_k^c$, formulated as:
\begin{equation}
    {\mathbf{P}}^c_k = \text{PI}'(\text{conca}(\hat{\mathbf{x}}_k,  \tilde{\mathbf{x}}_k)).
\end{equation}
This allows us to define the learned reverse 
process, which conditions on $\mathbf{P}_k^c$ 
instead of $\mathbf{P}_k^{h,0}$. $\mathbf{P}^{h,{t-1}}_k$ is updated by setting the variance to $1-\alpha^t$:
\begin{equation} \hspace{-4mm}
\scalebox{0.92}{$
        \mathbf{P}^{h,{t-1}}_k \!=\! \frac{1}{\sqrt{\alpha^t}}(\mathbf{P}^{h,{t}}_k \!- \frac{1-\alpha^t}{\sqrt{1-\bar{\alpha}^t}}\boldsymbol{\epsilon}_{\theta}(\mathbf{P}^{h,{t}}_k, \mathbf{P}^c_k, t))
       \! +\! \sqrt{1-\alpha^t}\boldsymbol{\epsilon}^t,$}
\end{equation} 
where $\boldsymbol{\epsilon}^t \sim \mathcal{N}(0,\mathbf{I})$. After $T$ denoising steps, DR-LDM outputs the reconstructed prior  
$\hat{\mathbf{P}}^h_k$, which is then used to guide OCFormer for fine-detail recovery.
In practice, only a few steps are required to produce a compact and high-quality prior, significantly reducing computational cost.

\noindent \textbf{Optimization}. To align the predicted and ground-truth priors, we define a diffusion consistency loss:
\begin{equation}
    \mathcal{L}_{Diff} = \|\hat{\mathbf{P}}^h_k-{\mathbf{P}}^h_k\|_1. 
\end{equation}

For end-to-end optimization, we jointly train all components of UnfoldLDM with the total Phase II objective:
\begin{equation}
    \mathcal{L}^{II}_{Total} = \mathcal{L}_{Rec} +  \mathcal{L}_{ISDA} + \mathcal{L}_{Diff}.
\end{equation}

\vspace{-3mm}
\subsection{Inference}
\vspace{-1mm}
During inference, at the $k^{th}$ stage, UnfoldLDM takes the degradation-corrected estimates $(\hat{\mathbf{x}}_k, \tilde{\mathbf{x}}_k)$ from MGDA and employs $\text{PI}'$ to encode the conditional prior cue ${\mathbf{P}}^c_k$. DR-LDM then synthesizes a high-quality prior $\hat{\mathbf{P}}^h_k$ conditioned on ${\mathbf{P}}^c_k$, 
which guides OCFormer to produce the restored image $\mathbf{x}_k$. 
Across successive stages, MGDA progressively refines degradation estimation, while DR-LDM and OCFormer collaboratively enhance texture recovery. This synergy ensures that the final output is both \textit{degradation-free and visually rich}.

\vspace{-2mm}
\section{Experiment}
\vspace{-2mm}
\label{sec:experiment}

\noindent \textbf{Experimental setup}. Our UnfoldLDM is implemented in PyTorch on two RTX 5090 GPUs using the Adam optimizer with momentum terms of (0.9, 0.999). The initial learning rate is set to $2\times10^{-4}$ and decayed to $1\times10^{-6}$ following the cosine annealing~\cite{loshchilovstochastic}. The stage number $K$ is set as 3.
The vector length $C_p$ is 64. The timestep of the diffusion model $T$ is set as 3. 
For OCFormer, the number of transformer blocks at levels 1–4 is configured as [2,2,2,2].

\vspace{-2mm}
\subsection{Comparative Evaluation on Paired Benchmarks} 
\vspace{-1mm}
We evaluate UnfoldLDM on six BIR tasks using PSNR and SSIM, spanning single-type degradation (denoising, deblurring, and deraining) and compound degradation (low-light, underwater, and backlit enhancement). Please see the detailed dataset description in the supplementary materials.

\begin{figure*}[t!]
\begin{minipage}[c]{\textwidth}
  \setlength{\abovecaptionskip}{-0.4cm}
  \centering
    \includegraphics[width=1\linewidth]{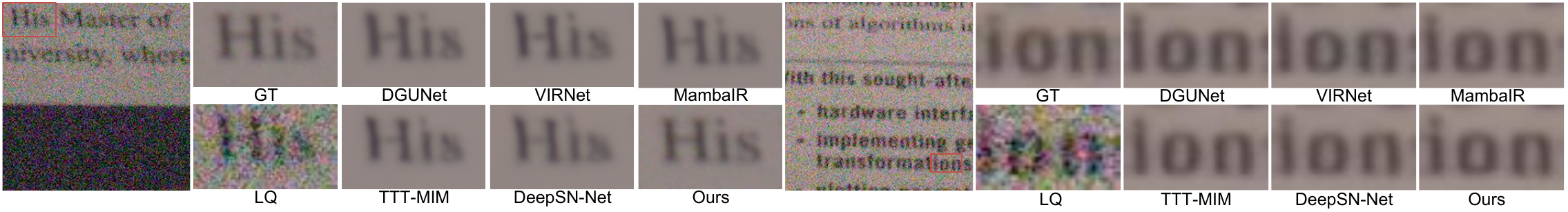}
    \newline
    \caption{Visualization of image denoising. Our method restores a sharper letter ``s'' in ``His'' (left) and a more accurate ``i'' in ``ion'' (right).}
    \label{Fig.denoise}
\end{minipage} \\
\begin{minipage}[c]{\textwidth}
  \setlength{\abovecaptionskip}{-0.4cm}
  \centering
    \includegraphics[width=1\linewidth]{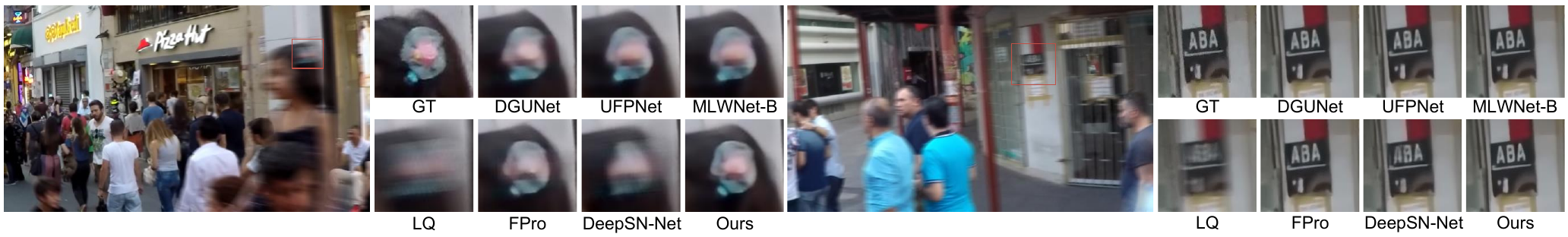}
    \newline
    \caption{Visualization of image deblurring. Our method resulted in a clearer hairpin (left) and a more accurate lettering ``A'' (right).}
    \label{Fig.deblur}
\end{minipage}\vspace{-0.4cm}
\end{figure*}

\noindent \textbf{Image denoising}.
Following DeepSN‑Net~\cite{deng2025deepsn}, we train UnfoldLDM on the training set of \textit{SIDD}~\cite{abdelhamed2018high} and evaluate it on the testing sets of \textit{SIDD} and \textit{DND}~\cite{plotz2017benchmarking}.
As reported in~\cref{table:denoising,Fig.denoise}, UnfoldLDM achieves SOTA performance, surpassing existing methods on both datasets.

\begin{table*}[t]
\begin{minipage}[c]{0.41\textwidth}
\centering
\setlength{\abovecaptionskip}{0cm}
\setlength{\belowcaptionskip}{0.05cm}
\caption{Results on denoising. The best two ones are in {\color[HTML]{FF0000}\textbf{red}} and {\color[HTML]{00B0F0}\textbf{blue}}.}\label{table:denoising}
\resizebox{\textwidth}{!}{
\setlength{\tabcolsep}{1.09mm}
\begin{tabular}{l|c|cc|cc}
\toprule
\multirow{2}{*}{Methods} & \multirow{2}{*}{Sources} & \multicolumn{2}{c|}{\textit{SIDD}} & \multicolumn{2}{c}{\textit{DND}} \\ \cline{3-6}
&  & \cellcolor{gray!40}PSNR~$\uparrow$ & \cellcolor{gray!40}SSIM~$\uparrow$       & \cellcolor{gray!40}PSNR~$\uparrow$ & \cellcolor{gray!40}SSIM~$\uparrow$     \\ \midrule
SDAP~\cite{pan2023random}                     & ICCV23                   & 37.53       & 0.936      & 38.56      & 0.940     \\
ADFNet~\cite{shen2023adaptive}                   & AAAI23                   & 39.63       & 0.958      & 39.87      & 0.955      \\
VIRNet~\cite{yue2024deep}                   & TPAMI24                   & 39.64       & 0.958      & 39.83      & 0.954      \\
MambaIR~\cite{guo2024mambair}                  & ECCV24                   &\color[HTML]{00B0F0} \textbf{39.89}       &\color[HTML]{00B0F0}  \textbf{0.960}       & \color[HTML]{00B0F0}\textbf{40.04}      &\color[HTML]{00B0F0} \textbf{0.956}      \\
TTT-MIM~\cite{mansour2024ttt}                  & ECCV24                   & 39.69       & ---          & 37.04      & ---          \\
DeepSN-Net~\cite{deng2025deepsn}               & TPAMI25                   & 39.79       & 0.958      & 39.92      & \color[HTML]{00B0F0}\textbf{0.956}      \\
DnLUT~\cite{yang2025dnlut}                    & CVPR25                   & ---           & 0.875      & 36.67      & 0.922      \\
\rowcolor{gray!10} UnfoldLDM                & Ours                     &\color[HTML]{FF0000} \textbf{40.23}       &\color[HTML]{FF0000} \textbf{0.965}      &    \color[HTML]{FF0000} \textbf{40.15}        &   \color[HTML]{FF0000} \textbf{0.964}       \\ \bottomrule 
\end{tabular}}
\end{minipage}
\begin{minipage}[c]{0.27\textwidth}
\centering
\setlength{\abovecaptionskip}{0cm}
\setlength{\belowcaptionskip}{0.05cm}
\caption{Results on the UIE task.}
\resizebox{\columnwidth}{!}{
\setlength{\tabcolsep}{0.8mm}
\begin{tabular}{l|c|cc}
\toprule
 && \multicolumn{2}{c}{\textit{UIEB}}\\ \cline{3-4}
\multirow{-2}{*}{Methods} &\multirow{-2}{*}{Sources} & \cellcolor{gray!40}PSNR~$\uparrow$ & \cellcolor{gray!40}SSIM~$\uparrow$ \\ \midrule
U-shape~\cite{peng2023u}& TIP23 & 22.91                                 &0.905 \\
PUGAN~\cite{cong2023pugan}& TIP23& { {23.05}} & 0.897   \\
ADP~\cite{zhou2023underwater}& IJCV23   & 22.90                                 & 0.892                          \\
NU2Net~\cite{guo2023underwater}   & AAAI23 & 22.38                                 & 0.903      \\
AST~\cite{Zhou_2024_CVPR}  & CVPR24 & 22.19 & 0.908 \\
MambaIR~\cite{guo2024mambair} &ECCV24 &22.60 &{ {0.916}}  \\
Reti-Diff~\cite{he2023reti}              & ICLR25        & {\color[HTML]{00B0F0} \textbf{24.12}} & {\color[HTML]{00B0F0} \textbf{0.910}} \\ 
\rowcolor{gray!10} UnfoldLDM & Ours &\color[HTML]{FF0000} \textbf{24.70} &\color[HTML]{FF0000} \textbf{0.947}
\\\bottomrule
\end{tabular}}
\label{table:Underwater} 
\end{minipage}
\begin{minipage}[c]{0.28\textwidth}
\centering
\setlength{\abovecaptionskip}{0cm}
\setlength{\belowcaptionskip}{0.05cm}
\caption{Results on the BIE task.}
\resizebox{\columnwidth}{!}{
\setlength{\tabcolsep}{0.6mm}
\begin{tabular}{l|c|cc}
\toprule
&  & \multicolumn{2}{c}{\textit{BAID}}                                                                                                                             \\ \cline{3-4}
\multirow{-2}{*}{Methods} & \multirow{-2}{*}{Sources} & \cellcolor{gray!40}PSNR~$\uparrow$ & \cellcolor{gray!40}SSIM~$\uparrow$ \\ \midrule
CLIP-LIT~\cite{liang2023iterative}                  & ICCV23                      & 21.13                                 & 0.853   \\
Diff-Retinex~\cite{yi2023diff}              & ICCV23                      &22.07 & 0.861     \\
DiffIR~\cite{xia2023diffir}                    & ICCV23                      & 21.10                                 & 0.835   \\
AST~\cite{Zhou_2024_CVPR}  & CVPR24 &22.61 & 0.851  \\
MambaIR~\cite{guo2024mambair} &ECCV24 &{ {23.07}} &{ {0.874}}  \\
RAVE~\cite{gaintseva2024rave} & ECCV24 & 21.26 & 0.872  \\
Reti-Diff~\cite{he2023reti}  & ICLR25                    & {\color[HTML]{00B0F0} \textbf{23.19}} & {\color[HTML]{00B0F0} \textbf{0.876}} \\
\rowcolor{gray!10} UnfoldLDM & Ours &\color[HTML]{FF0000} \textbf{24.97} &\color[HTML]{FF0000} \textbf{0.910} \\ 
\bottomrule
\end{tabular}}
\label{table:backlit}
\end{minipage} \\
\begin{minipage}[c]{0.435\textwidth}
\centering
\setlength{\abovecaptionskip}{0cm}
\caption{Results on deblurring.}\label{table:deblur}
\vspace{-4mm}
\resizebox{\textwidth}{!}{
\setlength{\tabcolsep}{1mm}
\begin{tabular}{l|c|cc|cc}
\toprule
\multirow{2}{*}{Methods} & \multirow{2}{*}{Sources} & \multicolumn{2}{c|}{\textit{GoPro}} & \multicolumn{2}{c}{\textit{HIDE}} \\ \cline{3-6}
& & \cellcolor{gray!40}PSNR~$\uparrow$  & \cellcolor{gray!40}SSIM~$\uparrow$       & \cellcolor{gray!40}PSNR~$\uparrow$  & \cellcolor{gray!40}SSIM~$\uparrow$      \\ \midrule
UFPNet~\cite{fang2023self}                   & CVPR23                   & 34.06       & 0.968       & 31.74       & 0.947      \\
FFTformer~\cite{kong2023efficient}                & CVPR23                   & 34.21       & 0.968       & 31.62       & 0.946      \\
MLWNet-B~\cite{gao2024efficient}                 & CVPR24                   & 33.83       & 0.968       & 31.06       & 0.932      \\
MISC Filter~\cite{liu2024motion}              & CVPR24                   & 34.10       &\color[HTML]{00B0F0} \textbf{0.969}      & 31.66       & 0.946      \\
FPro~\cite{zhou2024seeing}                     & ECCV24                   & 33.05       & 0.961       & 30.63       & 0.936      \\
DeepSN-Net~\cite{deng2025deepsn}               & TPAMI25                   & 32.83       & 0.960        & 31.14       & 0.941      \\
MDT~\cite{chen2025polarization}                      & CVPR25                   & \color[HTML]{00B0F0} \textbf{34.26}       &\color[HTML]{00B0F0} \textbf{0.969}       &\color[HTML]{00B0F0} \textbf{31.84}       & \color[HTML]{FF0000} \textbf{0.948}      \\
\rowcolor{gray!10} UnfoldLDM                & Ours                     &\color[HTML]{FF0000} \textbf{34.32}       &\color[HTML]{FF0000} \textbf{0.970}       & \color[HTML]{FF0000} \textbf{31.85}       &\color[HTML]{FF0000} \textbf{0.948}     \\ \bottomrule
\end{tabular}}
\end{minipage}
\begin{minipage}[c]{0.552\textwidth}
\centering
\setlength{\abovecaptionskip}{0cm}
\caption{Results on the LLIE task. 
}\label{table:low-light}
\vspace{-4mm}
\resizebox{\textwidth}{!}{
\setlength{\tabcolsep}{1mm}
\begin{tabular}{l|c|cc|cc|cc}
\toprule
&  & \multicolumn{2}{c|}{\textit{LOL-v1}}& \multicolumn{2}{c|}{\textit{LOL-v2-real}}& \multicolumn{2}{c}{\textit{LOL-v2-synthetic}}\\ \cline{3-8}
\multirow{-2}{*}{Methods} & \multirow{-2}{*}{Sources} 
& \cellcolor{gray!40}PSNR~$\uparrow$  & \cellcolor{gray!40}SSIM~$\uparrow$  
& \cellcolor{gray!40}PSNR~$\uparrow$  & \cellcolor{gray!40}SSIM~$\uparrow$  
& \cellcolor{gray!40}PSNR~$\uparrow$  & \cellcolor{gray!40}SSIM~$\uparrow$  \\ \midrule
Diff-Retinex~\cite{yi2023diff} & ICCV23 & 21.98 & 0.852 & 20.17 & 0.826 & 24.30 & 0.921 \\
CUE~\cite{zheng2023empowering} & ICCV23 & 21.86 & 0.841 & 21.19 & 0.829 & 24.41 & 0.917 \\
GSAD~\cite{jinhui2023global} & NIPS23 & 23.23 & 0.852 & 20.19 & 0.847 &24.22 & 0.927 \\
AST~\cite{Zhou_2024_CVPR} & CVPR24 & 21.09 & 0.858 & 21.68 & 0.856  & 22.25 & 0.927 \\
MambaIR~\cite{guo2024mambair} & ECCV24 & 22.23 & 0.863 & 21.15 & 0.857  & 25.75 & 0.937 \\
Reti-Diff~\cite{he2023reti} & ICLR25 & {\color[HTML]{00B0F0}\textbf{25.35}} & 0.866 & 22.97 & 0.858 & {\color[HTML]{00B0F0}\textbf{27.53}} &\color[HTML]{00B0F0} \textbf{0.951}  \\
CIDNet~\cite{yan2024you} & CVPR25 & 23.50 & {\color[HTML]{00B0F0}\textbf{0.900}}  & {\color[HTML]{FF0000}\textbf{24.11}} & \color[HTML]{00B0F0} \textbf{0.871} & 25.71 & 0.942 \\
\rowcolor{gray!10} UnfoldLDM & Ours & {\color[HTML]{FF0000}\textbf{25.58}} &\color[HTML]{FF0000} \textbf{0.912} & \color[HTML]{00B0F0} \textbf{23.88} &\color[HTML]{FF0000} \textbf{0.889} &\color[HTML]{FF0000} \textbf{27.92} &\color[HTML]{FF0000} \textbf{0.957}
\\ \bottomrule
\end{tabular}}
\end{minipage}
\vspace{-6mm}
\end{table*}

\noindent \textbf{Image deblurring}. 
Following \cite{liu2024motion,deng2025deepsn}, we train all models on the \textit{GoPro} dataset \cite{nah2017deep} and evaluate them on the test sets of \textit{GoPro} and \textit{HIDE} \cite{shen2019human}.
As reported in \cref{table:deblur,Fig.deblur}, our UnfoldLDM achieves a leading place qualitatively and quantitatively, consistently outperforming existing cutting‑edge approaches.

\begin{figure*}[t!]
  \setlength{\abovecaptionskip}{-0.4cm}
  \centering
    \includegraphics[width=\linewidth]{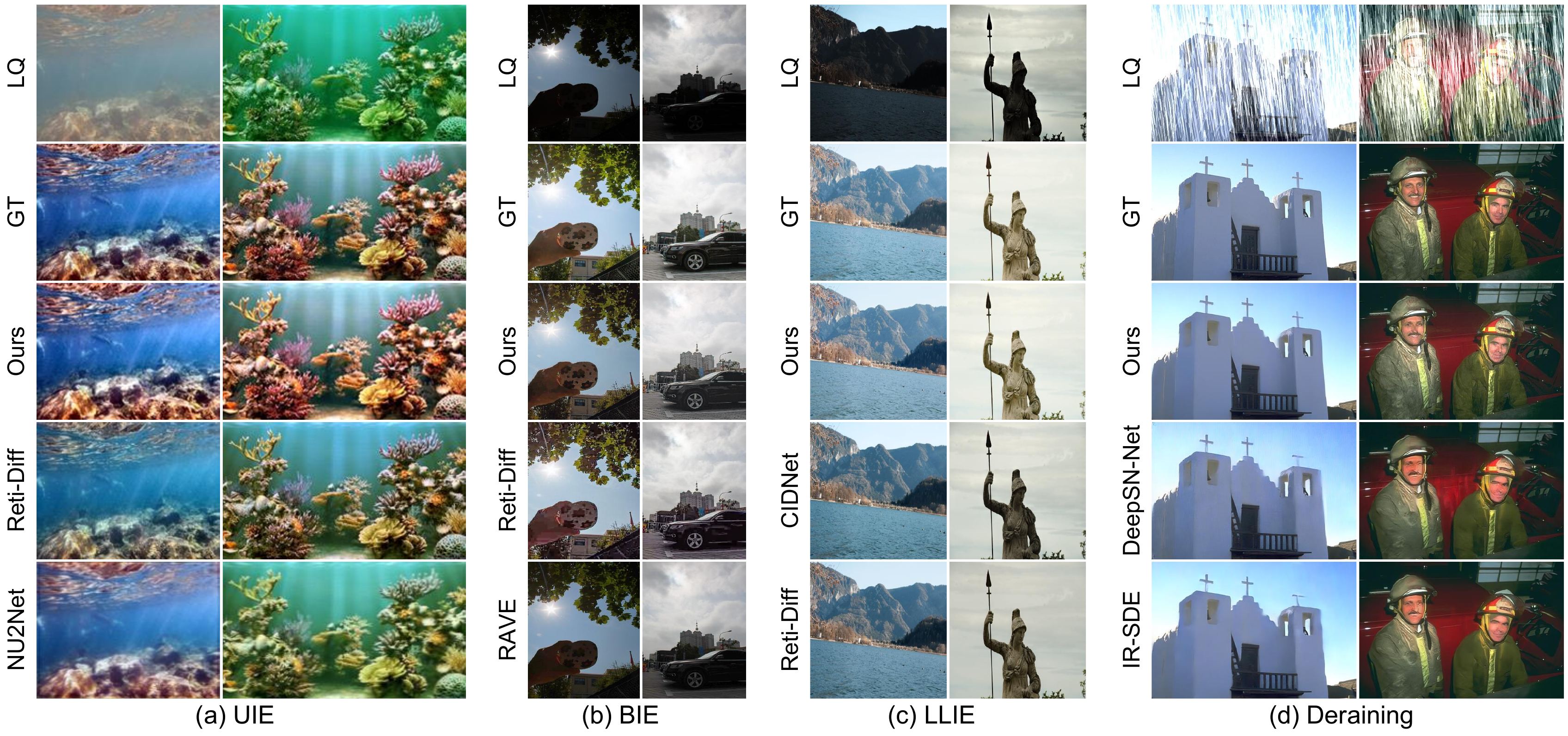}
    \newline
    \caption{Visualization on UIE, BIE, LLIE, and Deraining. Only SOTAs are selected.
    }
    \label{Fig.LOL_backlit_underwater}
\vspace{-0.4cm}
\end{figure*}

\noindent \textbf{Underwater image enhancement}. 
Following Reti‑Diff~\cite{he2023reti}, we evaluate our method on \textit{UIEB}~\cite{li2019underwater}.
As shown in~\cref{table:Underwater}, UnfoldLDM surpasses the second‑best approach, Reti‑Diff, by $3.24\%$ on average.
Together with the results shown in~\cref{Fig.LOL_backlit_underwater} with superior color correction, these findings verify our effectiveness.

\noindent \textbf{Backlit image enhancement}. Following CLIP‑LIT \cite{liang2023iterative}, we train and evaluate our model on \textit{BAID} \cite{lv2022backlitnet}.
As shown in~\cref{table:backlit}, our method outperforms the second‑best approach, Reti‑Diff, by $5.78\%$. Also, results shown in~\cref{Fig.LOL_backlit_underwater} indicates our superiority in anti-glare of the sun (left image), even surpassing the GT.

\begin{table*}[t]
\begin{minipage}[c]{\textwidth}
\centering
\setlength{\abovecaptionskip}{0cm}
\caption{Results on deraining (Resolution: $256\times 256$).
}\label{table:deraining}
\vspace{-4mm}
\resizebox{\textwidth}{!}{
\setlength{\tabcolsep}{1.6mm}
\begin{tabular}{l|c|cc|cc|cc|cc|cc|cc}
\toprule
&  &Para.  & FLOPs &\multicolumn{2}{c|}{\textit{Rain100L}} & \multicolumn{2}{c|}{\textit{Rain100H}}                                                     & \multicolumn{2}{c|}{\textit{Test100}}                                                   & \multicolumn{2}{c|}{\textit{Test1200}}                                                  & \multicolumn{2}{c}{\textit{Test2800}}                                                  \\ \cline{5-14}
\multirow{-2}{*}{Methods} & \multirow{-2}{*}{Sources} & (M) & (G) & \cellcolor{gray!40}PSNR~$\uparrow$  & \cellcolor{gray!40}SSIM~$\uparrow$ & \cellcolor{gray!40}PSNR~$\uparrow$  & \cellcolor{gray!40}SSIM~$\uparrow$& \cellcolor{gray!40}PSNR~$\uparrow$  & \cellcolor{gray!40}SSIM~$\uparrow$& \cellcolor{gray!40}PSNR~$\uparrow$  & \cellcolor{gray!40}SSIM~$\uparrow$& \cellcolor{gray!40}PSNR~$\uparrow$  & \cellcolor{gray!40}SSIM~$\uparrow$\\ \midrule
DGUNet~\cite{mou2022deep} & CVPR22 & 69.57 & 2335.46 & 37.42 & 0.969 & 30.66 & 0.891 & 30.32 & 0.899 &33.23 & 0.920 & 33.68 & 0.938  \\
IR-SDE~\cite{luo2023image}                    & ICML23      &  135.30  &    469.23            & 38.30                                 & {\color[HTML]{00B0F0} \textbf{0.980}} & 31.65                                 & 0.904                                 & ---                                     & ---                                     & ---                                     & ---                                     & 30.42                                 & 0.891                                 \\
MambaIR~\cite{guo2024mambair}                   & ECCV24      & 31.51  &  363.84                        & 38.78                                 & 0.977                                 & 30.62                                 & 0.893                                 & ---                                     & ---                                     & 32.56                                 & 0.923                                 & 33.58                                 & 0.927                                 \\
PRISM~\cite{xue2025prism}                     & ArXiv25     & ---  &    ---                      & 36.88                                 & 0.966                                 & 30.06                                 & 0.889                                 & 30.29                                 & 0.900                                 & 32.56                                 & 0.913                                 & 33.73                                 & 0.939                                 \\
DiNAT-IR~\cite{liu2025dinat}                  & ArXiv25     &  ---  &    ---                      & 38.93                                 & 0.977                                 & 31.26                                 & 0.903                                 & 31.22                                 & {\color[HTML]{00B0F0} \textbf{0.920}} & 32.31                                 & 0.923                                 & 33.91                                 & 0.943                                 \\
VMambaIR~\cite{shi2025vmambair}                  & TCSVT25      & 25.17   &  537.26                       & {\color[HTML]{00B0F0} \textbf{39.09}} & 0.979                                 & 31.66                                 & {\color[HTML]{00B0F0} \textbf{0.909}} & ---                                     & ---                                     & 33.33                                 & 0.926                                 & \color[HTML]{00B0F0} \textbf{34.01}                                 & {\color[HTML]{00B0F0} \textbf{0.944}} \\
DeepSN-Net~\cite{deng2025deepsn}                & TPAMI25     &  22.72  & 364.13                        & 38.59                                 & 0.975                                 & {\color[HTML]{00B0F0} \textbf{31.81}} & 0.904                                 &\color[HTML]{00B0F0} \textbf{31.60}                                 & {\color[HTML]{00B0F0} \textbf{0.920}} & {\color[HTML]{00B0F0} \textbf{33.45}} & {\color[HTML]{00B0F0} \textbf{0.931}} &\color[HTML]{00B0F0} \textbf{34.01}                                 & 0.942                                 \\
\rowcolor{gray!10} UnfoldLDM                 & Ours         &  23.76  &  87.33                       & {\color[HTML]{FF0000} \textbf{39.56}} & {\color[HTML]{FF0000} \textbf{0.983}} & {\color[HTML]{FF0000} \textbf{32.30}} & {\color[HTML]{FF0000} \textbf{0.914}} & {\color[HTML]{FF0000} \textbf{32.69}} & {\color[HTML]{FF0000} \textbf{0.928}} & {\color[HTML]{FF0000} \textbf{34.28}} & {\color[HTML]{FF0000} \textbf{0.937}} & {\color[HTML]{FF0000} \textbf{34.30}} & {\color[HTML]{FF0000} \textbf{0.948}} \\ \bottomrule
\end{tabular}} 
\end{minipage}\\
\begin{minipage}[c]{0.38\textwidth}
\centering
\setlength{\abovecaptionskip}{0cm}
\caption{Results on the real-world IDIR task.}\label{table:realIDIR} \vspace{-4mm}
\resizebox{\columnwidth}{!}{
\setlength{\tabcolsep}{1mm}
\begin{tabular}{l|cc|cc|cc}
\toprule
\multirow{2}{*}{Methods}& \multicolumn{2}{c|}{\textit{DICM}} & \multicolumn{2}{c|}{\textit{LIME}} & \multicolumn{2}{c}{\textit{MEF}}  \\
&\cellcolor{gray!40}PI~$\downarrow$       & \cellcolor{gray!40}NIQE~$\downarrow$              & \cellcolor{gray!40}PI~$\downarrow$       & \cellcolor{gray!40}NIQE~$\downarrow$              & \cellcolor{gray!40}PI~$\downarrow$      & \cellcolor{gray!40}NIQE~$\downarrow$                 \\ \midrule
GDP~\cite{fei2023generative} & {3.552}  & 4.358&   4.115 & 4.891 &  3.694  & 4.609  \\
Reti-Diff~\cite{he2023reti} &\color[HTML]{00B0F0} \textbf{2.351}     &\color[HTML]{00B0F0} \textbf{3.255}    & \color[HTML]{00B0F0} \textbf{2.837}     & \color[HTML]{00B0F0} \textbf{3.693}    &  {3.308} & {3.792}   \\ 
CIDNet~\cite{yan2024you}  &
3.045 & 3.796 & 3.146 & 4.132 &
\color[HTML]{00B0F0}\textbf{2.683} & 3.568   \\
UnfoldIR~\cite{he2025unfoldir} & {2.952} & 3.381 & 3.085 & 4.099 & 2.722 & \color[HTML]{00B0F0}\textbf{3.387} \\
\rowcolor{gray!10} UnfoldLDM & \color[HTML]{FF0000} \textbf{2.272} & \color[HTML]{FF0000} \textbf{3.122} & \color[HTML]{FF0000} \textbf{2.265} & \color[HTML]{FF0000} \textbf{3.336} & \color[HTML]{FF0000} \textbf{2.533} & \color[HTML]{FF0000} \textbf{3.152} \\
\bottomrule
\end{tabular}} 
\end{minipage}
\begin{minipage}[c]{0.6\textwidth}
\centering
\setlength{\abovecaptionskip}{0cm}
\caption{Results on blind image super-resolution (Resolution: $512\times 512$, GPU: H200). 
}\label{table:super-resolution}
\vspace{-4mm}
\resizebox{\columnwidth}{!}{
		\setlength{\tabcolsep}{1.2mm}
\begin{tabular}{l|cccc|cccc|c}
\toprule
\multirow{2}{*}{Methods} & \multicolumn{4}{c|}{\textit{RealSR}} & \multicolumn{4}{c|}{\textit{DRealSR}} & Time \\ \cline{2-9}
&\cellcolor{gray!40} PSNR$\uparrow$ &\cellcolor{gray!40}LPIPS$\downarrow$ & \cellcolor{gray!40}NIQE$\downarrow$ & \cellcolor{gray!40}MANIQA$\uparrow$ & \cellcolor{gray!40}PSNR$\uparrow$ & \cellcolor{gray!40}LPIPS$\downarrow$ & \cellcolor{gray!40}NIQE$\downarrow$ & \cellcolor{gray!40}MANIQA$\uparrow$ & (ms)  \\
\midrule
OSEDiff~\cite{wu2024one} & 25.15 & 0.292 & 5.648 & 0.633 & 27.92 &\color[HTML]{00B0F0} \textbf{0.297} & 6.490 & 0.590 &\color[HTML]{00B0F0} \textbf{63.85} \\
SinSR~\cite{wang2024sinsr} &\color[HTML]{00B0F0} \textbf{26.23} & 0.322 & 6.306 & 0.542 & 28.38 & 0.368 & 6.982 & 0.493 & 68.32   \\
TSD-SR~\cite{dong2025tsd} & 24.75 & 0.281 &\color[HTML]{00B0F0} \textbf{5.116} & 0.630 & 27.39 & 0.298 & \color[HTML]{00B0F0} \textbf{5.881} & 0.571 & 72.58 \\
FlowSR~\cite{xu2025fast} & 25.54 &\color[HTML]{00B0F0} \textbf{0.272} & 5.286 & \color[HTML]{00B0F0}\textbf{0.649} &\color[HTML]{FF0000} \textbf{28.50} & 0.298 & 6.135 &\color[HTML]{00B0F0}  \textbf{0.617} & --- \\ 
\rowcolor{gray!10}  {UnfoldLDM} & \color[HTML]{FF0000}\textbf{26.45} & \color[HTML]{FF0000}\textbf{0.270} & \color[HTML]{FF0000}\textbf{5.083} & \color[HTML]{FF0000}\textbf{0.653} &\color[HTML]{00B0F0} \textbf{28.47} &\color[HTML]{FF0000} \textbf{0.288} &\color[HTML]{FF0000} \textbf{5.851} &\color[HTML]{FF0000} \textbf{0.622} & \color[HTML]{FF0000}  \textbf{32.63} \\
\bottomrule
\end{tabular}}
\end{minipage}\vspace{-7mm}
\end{table*}

\noindent \textbf{Low-light image enhancement}. Following Reti-Diff \cite{he2023reti}, we evaluate our method on \textit{LOL‑v1} \cite{wei2018deep}, \textit{LOL‑v2‑real} \cite{yang2021sparse}, and \textit{LOL‑v2‑syn} \cite{yang2021sparse}. As shown in \cref{table:low-light}, UnfoldLDM achieves SOTA performance, outperforming the second‑best method (Reti‑Diff) and the third‑best method (CIDNet) by $2.36\%$ and $3.32\%$. Visualization presented in~\cref{Fig.LOL_backlit_underwater} also verify our superiority in visual fidelity.

\noindent \textbf{Image deraining}. Following DeepSN‑Net \cite{deng2025deepsn}, we evaluate performance on five datasets: \textit{Rain100H} \cite{yang2017deep}, \textit{Rain100L} \cite{yang2017deep}, \textit{Test100} \cite{zhang2019image}, \textit{Test2800} \cite{fu2017removing}, and \textit{Test1200} \cite{zhang2018density}.
As reported in \cref{table:deraining}, our method, with comparable efficiency, achieves SOTA results across all metrics and datasets.
Results in~\cref{Fig.LOL_backlit_underwater} demonstrates our effectiveness in restoring visual fidelity under severe raindrop occlusion.

\vspace{-2mm}
\subsection{Comparative Evaluation on Complex Real-world Degradations} 
\vspace{-1mm}

\noindent \textbf{Real-world illumination degradation image restoration}. Three real-world IDIR tasks, DICM~\cite{lee2013contrast}, LIME~\cite{guo2016lime}, and MEF~\cite{wang2013naturalness}, are selected for evaluation. Following~\cite{he2025unfoldir}, the model pretrained on \textit{LOL-v2-syn} are employed for inference with two metrics selected: PI~\cite{blau20182018} and NIQE~\cite{mittal2012making} (lower values indicate better results). As shown in~\cref{table:realIDIR}, our method outperforms existing methods. 

\noindent \textbf{Image super-resolution}. We evaluate UnfoldLDM on blind SR using RealSR~\cite{cai2019toward} and DRealSR~\cite{wei2020component} with three extra perception metrics: LPIPS~\cite{zhang2018unreasonable}, NIQE~\cite{zhang2015feature}, MANIQA~\cite{yang2022maniqa}. Following OSEDiff~\cite{wu2024one}, we train on LSDIR~\cite{li2023lsdir} and the first 10K face images from FFHQ~\cite{karras2019style} with the Real-ESRGAN degradation pipeline~\cite{wang2021real}. As presented in~\cref{table:super-resolution}, UnfoldLDM achieves the best trade-off between fidelity and perception, while being 2× faster than existing SOTAs. 

\begin{table*}[t!]
\begin{minipage}{\textwidth}
\centering
\setlength{\abovecaptionskip}{0cm}
\setlength{\belowcaptionskip}{0.05cm}
\caption{Ablation study of our UnfoldLDM.}\label{table:Ablation}
\resizebox{\textwidth}{!}{
\setlength{\tabcolsep}{1mm}
\begin{tabular}{l|c|ccccccc|ccc|cc|c}
\toprule
\multirow{2}{*}{Datasets}        & \multirow{2}{*}{Metrics} & \multicolumn{7}{c|}{MGDA} &  \multicolumn{3}{c|}{OCFormer}&  \multicolumn{2}{c|}{Training}& \cellcolor{gray!10} UnfoldLDM \\ \cline{3-14}
& & Retinex& w/o $\hat{\mathbf{x}}_k$ & w/o $\tilde{\mathbf{x}}_k$ & w/o $VSS(\bigcdot)$ & $\mathcal{D}_1(\bigcdot)$ & $\mathcal{D}_2(\bigcdot)$ & w/o $\mathcal{L}_{ISDA}$ & w/o DRA & w/o PDR  & w/o DR-LDM & One phase & w/o joint &\cellcolor{gray!10}  Ours      \\ \midrule
\multirow{2}{*}{\textit{L-v2-r}} & PSNR        & 23.27             & 23.25                         & 22.71                           & 23.06      & 22.87                  & 23.15                  & 22.76    & 23.26   & 22.39    & 22.08  & 21.07 &  22.86    &\cellcolor{gray!10}  \textbf{23.88}     \\
 & SSIM        & 0.878             & 0.883                         & 0.858                           & 0.879      & 0.855                  & 0.876                  & 0.872    & 0.870   & 0.858    & 0.853    &  0.835 &  0.868      &\cellcolor{gray!10}  \textbf{0.889}     \\ \midrule
\multirow{2}{*}{\textit{L-v2-s}} & PSNR      & 27.73               & 27.69                         & 26.61                           & 27.52      & 26.65                  & 27.33                  & 26.27    & 27.38   & 25.96    & 25.27    & 25.88 & 26.23       &\cellcolor{gray!10}  \textbf{27.92}     \\
& SSIM      & 0.949               & 0.952                         & 0.933                           & 0.952      & 0.940                  & 0.949                  & 0.950    & 0.949   & 0.941    & 0.929   & 0.861 &   0.943      &\cellcolor{gray!10}  \textbf{0.957}  \\ \bottomrule   
\end{tabular}}
\end{minipage}\\
\begin{minipage}{\textwidth}
\centering
\setlength{\abovecaptionskip}{0cm}
\setlength{\belowcaptionskip}{0.05cm}
\caption{Parameter analysis of the stage number $K$, timestep $T$, and vector length $C_P$. We also include a baseline for prior estimation (vector length: $C^1_p$), where DR-LDM encodes only the low-quality input $\mathbf{y}$ as the prior and uses it to guide the network.
}
\label{table:ParaAnalysis}
\resizebox{\textwidth}{!}{
\setlength{\tabcolsep}{1.4mm}
\begin{tabular}{l|c|cccc|cccc|cccc|cccc} 
\toprule
\multirow{2}{*}{Datasets}        & \multirow{2}{*}{Metrics} & \multicolumn{4}{c|}{Stage number $K$} & \multicolumn{4}{c|}{Timestep $T$}     & \multicolumn{4}{c|}{Vector length $C_p$} & \multicolumn{4}{c}{Vector length $C^1_p$}        \\ \cline{3-18}
& & K=2   &\cellcolor{gray!10} K=3 & K=4   & K=5   & T=2   & \cellcolor{gray!10} T=3 & T=4   & T=5   & $C_p$=16 & $C_p$=32 & \cellcolor{gray!10} $C_p$=64 & $C_p$=128 & $C^1_p$=16 & $C^1_p$=32 & $C^1_p$=64 & $C^1_p$=128 \\ \midrule
\multirow{2}{*}{\textit{L-v2-r}} & PSNR                     & 23.03 & \cellcolor{gray!10} 23.88      & 24.17 & \textbf{24.82} & 22.43 & \cellcolor{gray!10} 23.88      & 23.60 & 23.52 & 22.17   & 23.17   & \cellcolor{gray!10} 23.88          & 23.63 & 21.16 & 22.08 & 22.27 & 22.07   \\
& SSIM                     & 0.872 & \cellcolor{gray!10} 0.889      & 0.893 & \textbf{0.901} & 0.861 & \cellcolor{gray!10} 0.889      & 0.888 & 0.883 & 0.867   & 0.880   & \cellcolor{gray!10} 0.889          & 0.888  &  0.845 & 0.856 & 0.863 & 0.855  \\ \midrule
\multirow{2}{*}{\textit{L-v2-s}} & PSNR                     & 27.25 & \cellcolor{gray!10} 27.92      & 28.38 & \textbf{28.83} & 26.05 & \cellcolor{gray!10} 27.92      & 28.03 & 28.15 & 26.83   & 27.28   &\cellcolor{gray!10}  27.92          & 28.25  &  25.14 & 25.83 & 26.35 & 25.61  \\
& SSIM                     & 0.948 &\cellcolor{gray!10}  0.957      & 0.963 & \textbf{0.964} & 0.943 &\cellcolor{gray!10}  0.957      & 0.956 & 0.958 & 0.942   & 0.950   &\cellcolor{gray!10}  0.957          & 0.961 & 0.913 & 0.920 & 0.932 & 0.922  \\ \bottomrule
\end{tabular}}
\end{minipage}\vspace{-3mm}
\end{table*}

\vspace{-2mm}
\subsection{Ablation Study}\vspace{-1mm}
We conduct ablation study on low-light image enhancement with \textit{LOL-v2-real} (\textit{L-v2-r}) and \textit{LOL-v2-synthetic} (\textit{L-v2-s}).

\begin{figure*}[t!]
\begin{minipage}[c]{0.49\textwidth}
\centering
  \setlength{\abovecaptionskip}{-0.0cm}
\includegraphics[width=\columnwidth]{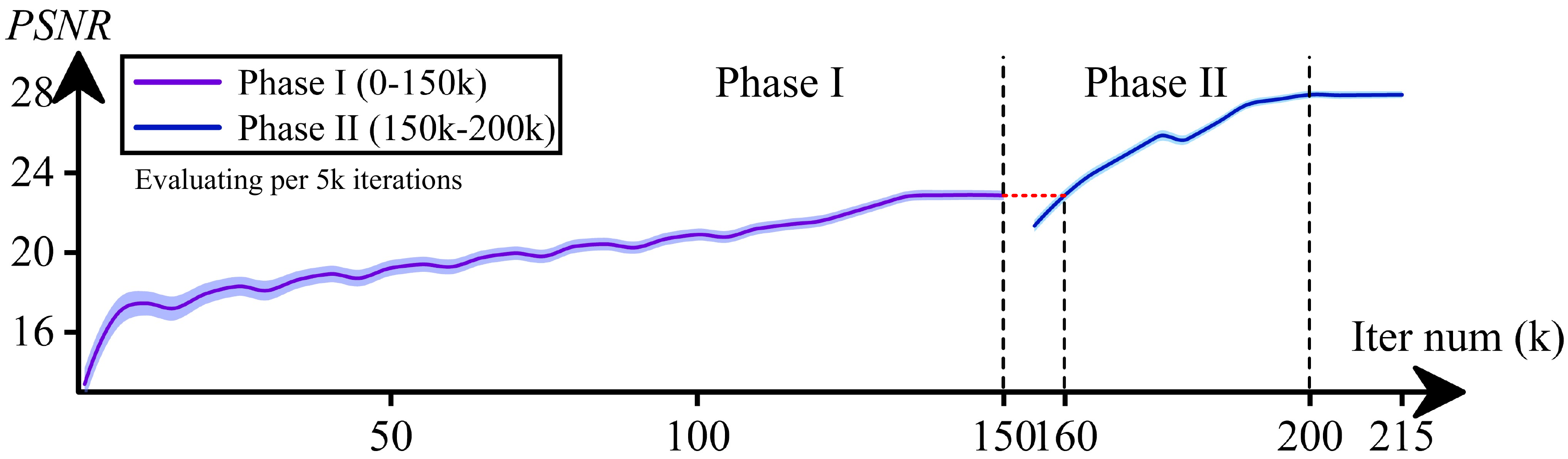}
\caption{Convergence analysis.
}
\label{fig:convergence}
\end{minipage}
\begin{minipage}[c]{0.49\textwidth}
\centering
  \setlength{\abovecaptionskip}{-0.0cm}
		\begin{subfigure}{0.185\textwidth}
		\centering
		\includegraphics[width=\textwidth]{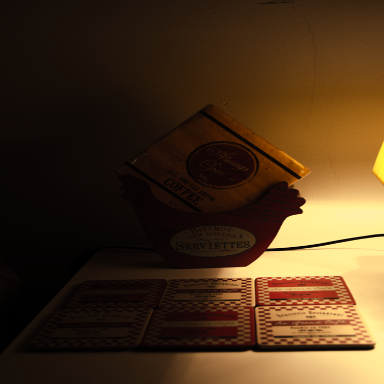}\vspace{-2pt}
		\caption*{ LQ }
	\end{subfigure}
	\begin{subfigure}{0.185\textwidth}  
		\centering 
		\includegraphics[width=\textwidth]{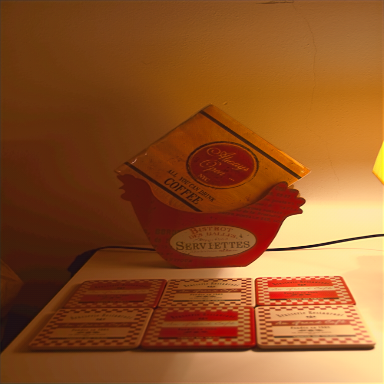}\vspace{-2pt}
		\caption*{ ${\mathbf{x}}_1$ }
	\end{subfigure}
	\begin{subfigure}{0.185\textwidth}  
		\centering 
		\includegraphics[width=\textwidth]{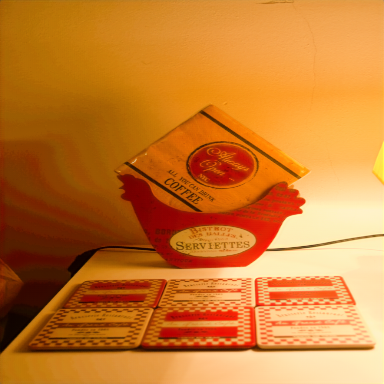}\vspace{-2pt}
		\caption*{${\mathbf{x}}_2$}
	\end{subfigure}
    	\begin{subfigure}{0.185\textwidth}  
		\centering 
		\includegraphics[width=\textwidth]{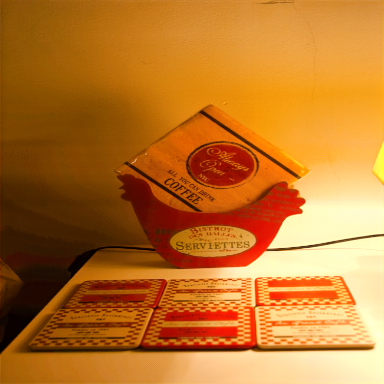}\vspace{-2pt}
		\caption*{${\mathbf{x}}_3$}
	\end{subfigure}
	\begin{subfigure}{0.185\textwidth}  
		\centering 
		\includegraphics[width=\textwidth]{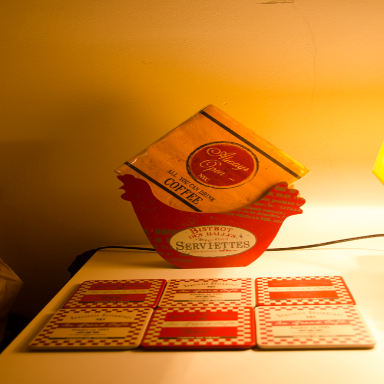}\vspace{-2pt}
		\caption*{GT}
	\end{subfigure}
	\caption{Stage-wise visualization with increased fidelity and details.
 }\label{fig:stagewise}
    \end{minipage}	\vspace{-6mm}
\end{figure*}

\noindent \textbf{Effect of MGDA}. As shown in~\cref{table:Ablation}, removing any key components ($\hat{\mathbf{x}}_k$, $\tilde{\mathbf{x}}_k$, $VSS(\bigcdot)$, and $\mathcal{L}_{ISDA}$) causes clear performance drops.
Replacing our adaptive degradation estimation (for unknown degradations) with the Retinex model \cite{wu2022uretinex} also degrades results, as Retinex relies on a pretrained network with limited capacity and cannot handle mixed degradations beyond illumination.
Substituting $\mathcal{D}_{\mathbf{M}}(\bigcdot)$ and $\mathcal{D}_{\mathbf{W}}(\bigcdot)$ with a direct network mapping ($\mathcal{D}_1(\bigcdot)$) or SVD-based approximation ($\mathcal{D}_2(\bigcdot)$) further confirms the advantage of our degradation modeling.

\begin{figure*}[t!]
\begin{minipage}[c]{0.49\textwidth}
\centering
  \setlength{\abovecaptionskip}{-0.0cm}
\includegraphics[width=\textwidth]{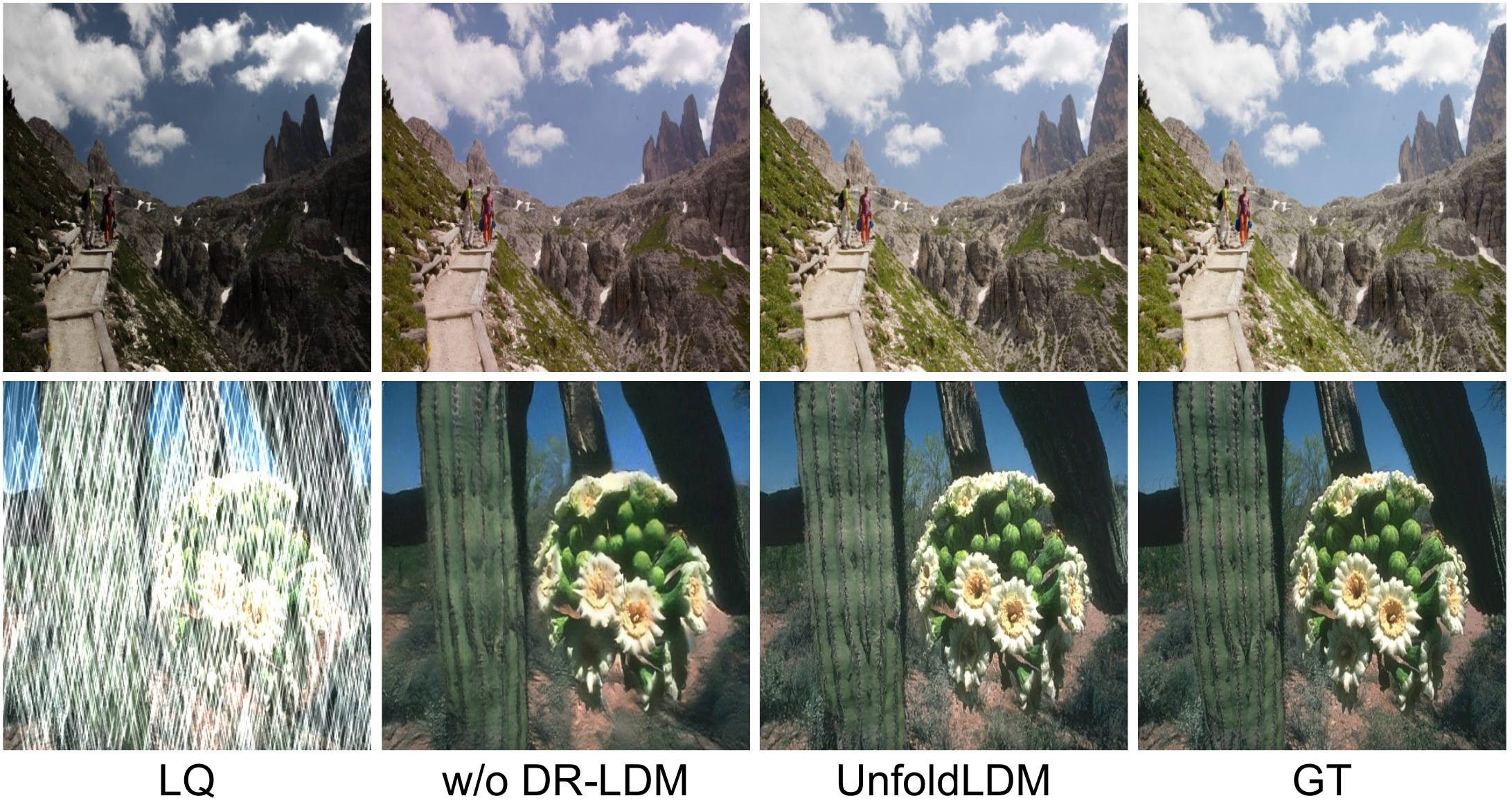}
	\caption{Effect of DR-LDM in visual fidelity and detail highlight.
 }
	\label{fig:VisualAblation}
\end{minipage}
\begin{minipage}[c]{0.49\textwidth}
\centering
  \setlength{\abovecaptionskip}{-0.0cm}
\includegraphics[width=\textwidth]{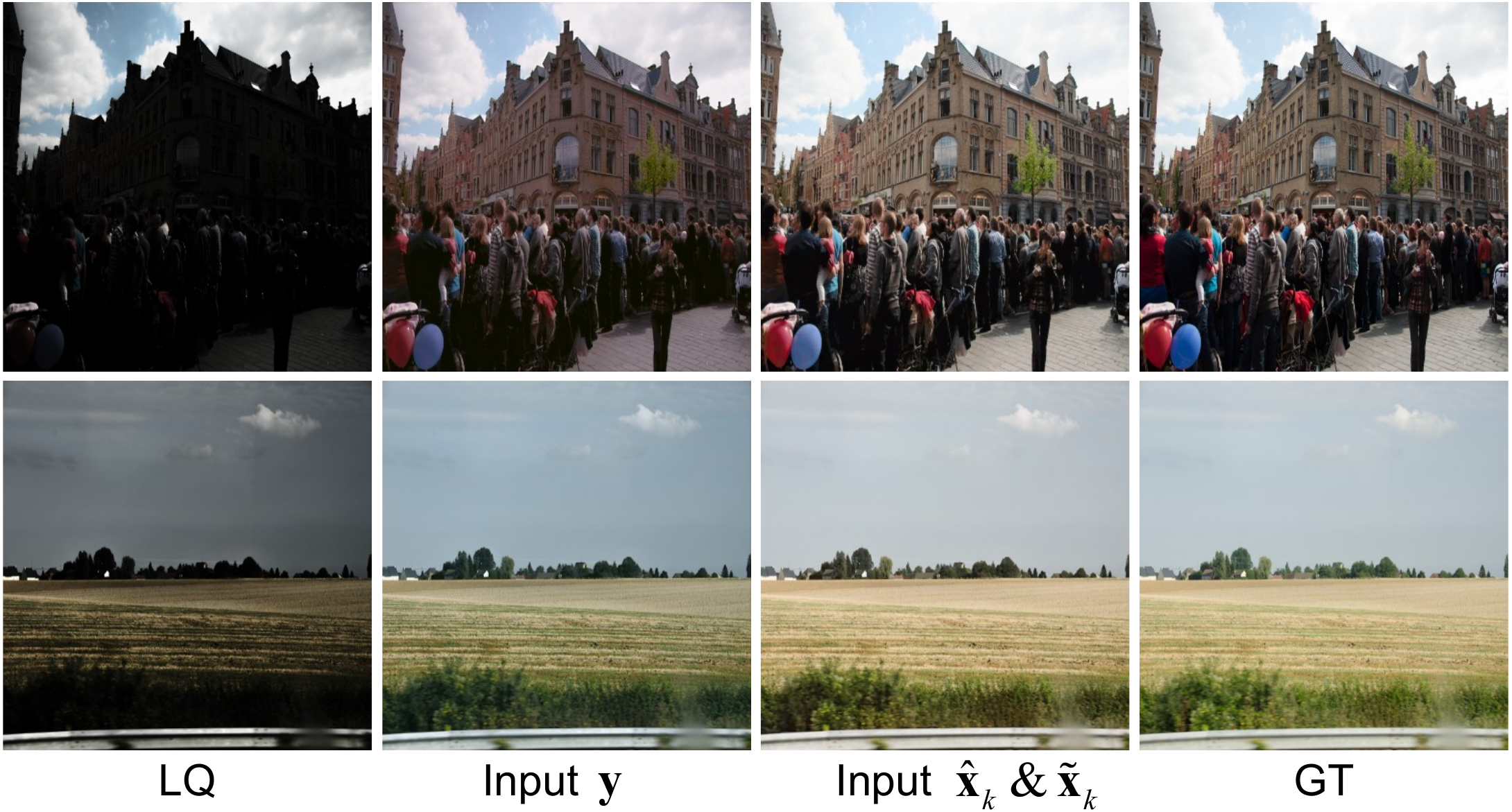} 
	\caption{Priors encoded by DR-LDM with different inputs.
 }
	\label{fig:EffectDR-LDM}
    \end{minipage}	\\
\begin{minipage}[c]{\textwidth}
\centering
		\begin{subfigure}{0.115\textwidth}
		\centering
		\includegraphics[width=\textwidth]{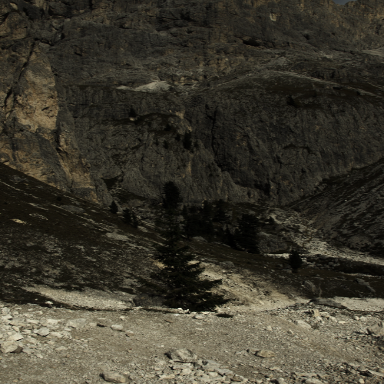}\vspace{-2pt}
		\caption{LQ}
	\end{subfigure}
	\begin{subfigure}{0.115\textwidth}  
		\centering 
		\includegraphics[width=\textwidth]{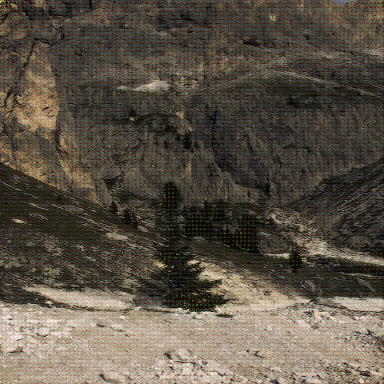}\vspace{-2pt}
		\caption{$\hat{\mathbf{x}}_2$}
	\end{subfigure}
	\begin{subfigure}{0.115\textwidth}  
		\centering 
		\includegraphics[width=\textwidth]{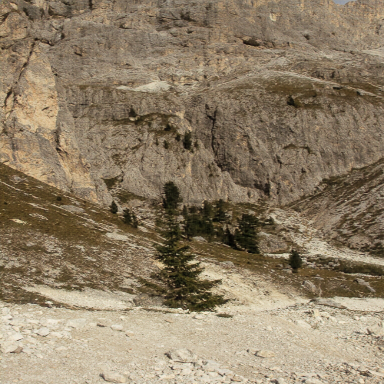}\vspace{-2pt}
		\caption{$\tilde{\mathbf{x}}_2$}
	\end{subfigure}
	\begin{subfigure}{0.115\textwidth}  
		\centering 
		\includegraphics[width=\textwidth]{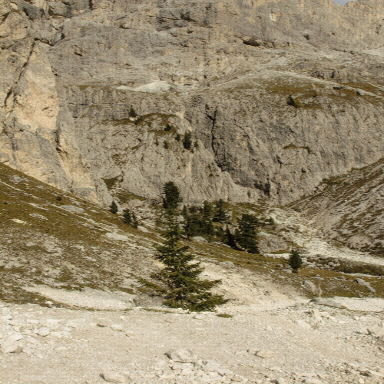}\vspace{-2pt}
		\caption{$\mathbf{x}_2$}
	\end{subfigure}
		\begin{subfigure}{0.115\textwidth}
		\centering
		\includegraphics[width=\textwidth]{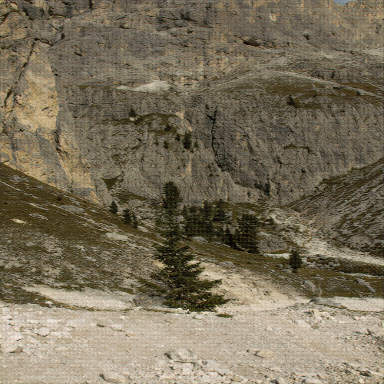}\vspace{-2pt}
		\caption{$\hat{\mathbf{x}}_3$}
	\end{subfigure}
	\begin{subfigure}{0.115\textwidth}  
		\centering 
		\includegraphics[width=\textwidth]{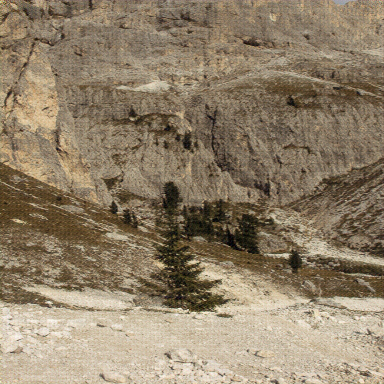}\vspace{-2pt}
		\caption{$\tilde{\mathbf{x}}_3$}
	\end{subfigure}
	\begin{subfigure}{0.115\textwidth}  
		\centering 
		\includegraphics[width=\textwidth]{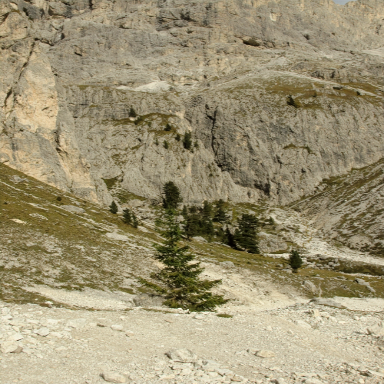}\vspace{-2pt}
		\caption{$\mathbf{x}_3$}
	\end{subfigure}
	\begin{subfigure}{0.115\textwidth}  
		\centering 
		\includegraphics[width=\textwidth]{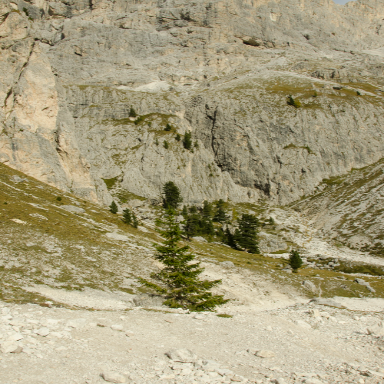}\vspace{-2pt}
		\caption{GT}
	\end{subfigure}\vspace{-3mm}
	\caption{Visualization at different stage, where the quality of $\hat{\mathbf{x}}_k$ and $\tilde{\mathbf{x}}_k$ increases with stage and jointly promote the quality of ${\mathbf{x}}_k$.
 }
	\label{fig:VisualizationStage}
    \end{minipage}
	\vspace{-3mm}
\end{figure*}
\begin{table}[t]
\centering
\caption{Efficiency comparison on \textit{LOL-v2-syn} (batch size: 1, resolution: $256 \times 256$, GPU: RTX 5090). UnfoldLDM-l is the lightweight version of UnfoldLDM.}
\label{table:efficiency}
\vspace{-2.5mm}
\resizebox{\columnwidth}{!}{
\setlength{\tabcolsep}{1.0mm}
\begin{tabular}{l|ccccc|ccccc}
\toprule
Methods & Para. (M) & FLOPs (G) & Memory (G) & Train (h) & Test (ms) 
& PSNR $\uparrow$ & SSIM $\uparrow$ & FID $\downarrow$ & BIQE $\uparrow$ & MANIQA $\uparrow$ \\
\midrule
Reti-Diff~\cite{he2023reti} & 26.11 & 87.63 & 11.55 & 30.89 &  13.73
& 27.53 & 0.951 & 13.26 & 15.77 & 0.455 \\
CIDNet~\cite{yan2024you} & 1.88 & 7.57 & 1.03 & 4.86 & 2.95 
& 25.71 & 0.942 & 18.60 & 15.87 & 0.439 \\
\rowcolor{gray!10} UnfoldLDM & 26.91 & 87.33 & 13.50
& 18.59 & 11.32 & \textbf{27.92} & \textbf{0.957} & \textbf{10.65} & \textbf{22.52} & \textbf{0.487} \\
\rowcolor{gray!10} UnfoldLDM-l & \textbf{1.06} & \textbf{2.12} & \textbf{0.56} 
& \textbf{2.02} & \textbf{1.37} & 26.36 & 0.950 & 15.03 & 16.70  & 0.453\\
\bottomrule
\end{tabular}}
\vspace{-6mm}
\end{table}
\begin{table*}[t]
\begin{minipage}[c]{0.3\textwidth}
\centering
\setlength{\abovecaptionskip}{0.05cm}
\caption{ User study. }
		\label{table:UserStudy} 
        \vspace{-4mm}
\resizebox{\columnwidth}{!}{
\setlength{\tabcolsep}{1.3mm}
\begin{tabular}{l|cccc}
\toprule
Methods &\multicolumn{1}{c}{\cellcolor{c2!50} \textit{L-v1}}           & \multicolumn{1}{c}{\cellcolor{c2!50} \textit{L-v2}}      & \multicolumn{1}{c}{\cellcolor{c2!50} \textit{UIEB}}       & \multicolumn{1}{c}{\cellcolor{c2!50} \textit{BAID}}   \\ \midrule
Uretinex~\cite{wu2022uretinex}  &  2.58    &   3.08  & ---     &  3.17   \\
CUE~\cite{zheng2023empowering}       &    2.50   &  2.92  & ---                                    & ---                                    \\
MambaIR~\cite{guo2024mambair}   &  3.42  &  4.08 & \color[HTML]{00B0F0} \textbf{3.67}   &  \color[HTML]{00B0F0} \textbf{3.67}     \\
Reti-Diff~\cite{he2023reti} & \color[HTML]{00B0F0} \textbf{3.58}  & \color[HTML]{00B0F0} \textbf{4.17}  & 3.50 & 3.58 \\
CIDNet~\cite{yan2024you}    & 3.17   & 3.92   & ---                                   & ---                                    \\
\rowcolor{gray!10} UnfoldLDM  & \color[HTML]{FF0000} \textbf{3.92}  & \color[HTML]{FF0000} \textbf{4.25}  & \color[HTML]{FF0000} \textbf{4.08}  & \color[HTML]{FF0000} \textbf{3.83}  \\ \bottomrule
\end{tabular}}
\end{minipage}
\begin{minipage}[c]{0.695\textwidth}
\centering
\setlength{\abovecaptionskip}{0.05cm}
\caption{Low-light image detection on \textit{ExDark}.
}
		\label{table:Detection} 
        \vspace{-4mm}
\resizebox{\columnwidth}{!}{
\setlength{\tabcolsep}{1.3mm}
\begin{tabular}{l|cccccccccccc|c}
\toprule
Methods (AP)& \cellcolor{c2!50}Bicycle  & \cellcolor{c2!50}Boat & \cellcolor{c2!50}Bottle  & \cellcolor{c2!50}Bus & \cellcolor{c2!50}Car & \cellcolor{c2!50}Cat  & \cellcolor{c2!50}Chair &\cellcolor{c2!50} Cup & \cellcolor{c2!50}Dog & \cellcolor{c2!50} Motor & \cellcolor{c2!50}People & \cellcolor{c2!50}Table  & \cellcolor{c2!50}Mean \\ \midrule
Baseline & 74.7                                 & 64.9                                 & 70.7                                 & 84.2                                 & 79.7                                 & 47.3                                 & 58.6                                 & 67.1                                 & 64.1                                 & 66.2                                 & 73.9                                 & 45.7                                 & 66.4                                 \\
SCI~\cite{ma2022toward}   & 73.4                                 & 68.0                                 & 69.5                                 & 86.2                                 & 74.5                                 & 63.1                                 & 59.5                                 & 61.0                                 & 67.3                                 & 63.9                                 & 73.2                                 & 47.3                                 & 67.2                                 \\
SNR-Net~\cite{xu2022snr}   & {{78.3}} & 74.2                                 & {{74.5}} & 89.6                                 & {{82.7}} & {{66.8}} & 66.3                                 & 62.5                                 & 74.7                                 & 63.1                                 & 73.3                                 & {{57.2}} & 71.9                                 \\
Reti-Diff~\cite{he2023reti} & {\color[HTML]{00B0F0} \textbf{82.0}} & {\color[HTML]{00B0F0} \textbf{77.9}} & {{76.4}} & {\color[HTML]{FF0000} \textbf{92.2}} & {\color[HTML]{FF0000} \textbf{83.3}} & {\color[HTML]{00B0F0} \textbf{69.6}} & {\color[HTML]{00B0F0} \textbf{67.4}} & {\color[HTML]{00B0F0} \textbf{74.4}} & {\color[HTML]{00B0F0} \textbf{75.5}} & {\color[HTML]{00B0F0} \textbf{74.3}} & {\color[HTML]{00B0F0} \textbf{78.3}} & {{57.9}} & {\color[HTML]{00B0F0} \textbf{75.8}}\\
CIDNet~\cite{yan2024you} & 81.8 & 77.6 & \color[HTML]{00B0F0} \textbf{77.2} & 85.8 & 77.3 & 68.1 & 65.5 & 73.6 & 74.7 & 70.2 & 71.0 & \color[HTML]{00B0F0} \textbf{60.3} & 73.6 \\
\rowcolor{gray!10} UnfoldLDM &\color[HTML]{FF0000} \textbf{88.2} &\color[HTML]{FF0000} \textbf{81.8} &\color[HTML]{FF0000} \textbf{77.7} &\color[HTML]{00B0F0} \textbf{90.6} & \color[HTML]{00B0F0}\textbf{82.8} &\color[HTML]{FF0000} \textbf{77.3} &\color[HTML]{FF0000} \textbf{81.2} &\color[HTML]{FF0000} \textbf{79.5} &\color[HTML]{FF0000} \textbf{82.7} &\color[HTML]{FF0000} \textbf{85.1} &\color[HTML]{FF0000} \textbf{79.8} &\color[HTML]{FF0000} \textbf{66.7} & \color[HTML]{FF0000}\textbf{81.1}
\\ \bottomrule
\end{tabular}}
\end{minipage}\\
\begin{minipage}[c]{\textwidth}
\centering
\setlength{\abovecaptionskip}{0cm}
\setlength{\belowcaptionskip}{0.05cm}
\caption{Generalization of UnfoldLDM, where we incorporate our DR-LDM with deep unfolding networks of different tasks and observe consistent performance gains.}\label{table:generalization}
\begin{subtable}[c]{0.32\textwidth}
\centering
\setlength{\abovecaptionskip}{0cm}
\setlength{\belowcaptionskip}{0.05cm}
\caption{Low-light enhancement}
\resizebox{\columnwidth}{!}{
\setlength{\tabcolsep}{0.47mm}
\begin{tabular}{l|c|cc}
\toprule
Datasets                         & Metrics & URetinex~\cite{wu2022uretinex} &\cellcolor{gray!10} URetinex+DR-LDM \\ \midrule
\multirow{2}{*}{\textit{L-v2-r}} & PSNR~$\uparrow$    & 20.44    & \cellcolor{gray!10}\textbf{21.93}           \\
                                 & SSIM~$\uparrow$    & 0.806    & \cellcolor{gray!10}\textbf{0.853}           \\
                                 \midrule
\multirow{2}{*}{\textit{L-v2-s}} & PSNR~$\uparrow$    & 24.73    &\cellcolor{gray!10}\textbf{26.08}          \\
                                 & SSIM~$\uparrow$    & 0.897    & \cellcolor{gray!10}\textbf{0.910}           \\
                                 \bottomrule         
\end{tabular}}
\label{table:LDMLLIE} 
\end{subtable}
\begin{subtable}[c]{0.305\textwidth}
\centering
\setlength{\abovecaptionskip}{0cm}
\setlength{\belowcaptionskip}{0.05cm}
\caption{Image fusion}
\resizebox{\columnwidth}{!}{
\setlength{\tabcolsep}{0.6mm}
\begin{tabular}{l|c|cc}
\toprule
Datasets                      & Metrics & IVF-Net~\cite{ju2022ivf} & \cellcolor{gray!10}IVF-Net+DR-LDM \\ \midrule
\multirow{2}{*}{\textit{TNO}} & EN~$\uparrow$      & 7.12    &\cellcolor{gray!10} \textbf{7.62}           \\
                              & NCC~$\uparrow$     & 0.806   &\cellcolor{gray!10} \textbf{0.825}          \\
                              \midrule
\multirow{2}{*}{\textit{INO}} & EN~$\uparrow$      & 7.54    &\cellcolor{gray!10} \textbf{8.13}           \\
                              & NCC~$\uparrow$     & 0.811   &\cellcolor{gray!10} \textbf{0.829}          \\
                              \bottomrule     
\end{tabular}}
\label{table:LDMIVF} 
\end{subtable}
\begin{subtable}[c]{0.353\textwidth}
\centering
\setlength{\abovecaptionskip}{0cm}
\setlength{\belowcaptionskip}{0.05cm}
\caption{Image deblurring}
\resizebox{\columnwidth}{!}{
\setlength{\tabcolsep}{0.6mm}
\begin{tabular}{l|c|cc}
\toprule
Datasets                        & Metrics & DeepSN-Net~\cite{deng2025deepsn} &\cellcolor{gray!10} DeepSN-Net+DR-LDM \\ \midrule
\multirow{2}{*}{\textit{GoPro}} & PSNR~$\uparrow$    & 32.83      &\cellcolor{gray!10} \textbf{33.76}       \\
& SSIM~$\uparrow$    & 0.960      &\cellcolor{gray!10} \textbf{0.966}       \\
                                \midrule
\multirow{2}{*}{\textit{HIDE}}  & PSNR~$\uparrow$    & 31.14      &\cellcolor{gray!10} \textbf{31.93}       \\
                                & SSIM~$\uparrow$    & 0.941      &\cellcolor{gray!10} \textbf{0.946}       \\
                                \bottomrule
\end{tabular}}
\label{table:LDMdeblur}
\end{subtable}
 \\
 \begin{subtable}[c]{0.34\textwidth}
\centering
\setlength{\abovecaptionskip}{0cm}
\setlength{\belowcaptionskip}{0.05cm}
\caption{Image deraining}
\resizebox{\columnwidth}{!}{
\setlength{\tabcolsep}{1.1mm}
\begin{tabular}{l|c|cc}
\toprule
Datasets                           & Metrics & DGUNet~\cite{mou2022deep} & \cellcolor{gray!10}DGUNet+DR-LDM \\ \midrule
\multirow{2}{*}{\textit{Rain100L}} & PSNR~$\uparrow$    & 38.25  &\cellcolor{gray!10} \textbf{38.87}   \\
& SSIM~$\uparrow$    & 0.974  &\cellcolor{gray!10} \textbf{0.978}   \\
\midrule
\multirow{2}{*}{\textit{Rain100H}} & PSNR~$\uparrow$    & 31.06  &\cellcolor{gray!10} \textbf{31.89}   \\
& SSIM~$\uparrow$    & 0.897  &\cellcolor{gray!10} \textbf{0.905}   \\
\bottomrule
\end{tabular}}
\label{table:LDMderain} 
\end{subtable}
 \begin{subtable}[c]{0.28\textwidth}
\centering
\setlength{\abovecaptionskip}{0cm}
\setlength{\belowcaptionskip}{0.05cm}
\caption{Salient object detection}
\resizebox{\columnwidth}{!}{
\setlength{\tabcolsep}{1.1mm}
\begin{tabular}{l|c|cc}\toprule
Datasets                       & Metrics & RUN~\cite{he2025run}   & \cellcolor{gray!10}RUN+DR-LDM \\ \midrule
\multirow{2}{*}{\textit{DUTS}} & $M$~$\downarrow$       & 0.022 &\cellcolor{gray!10}\textbf{0.021}      \\
& $F_\beta$~$\uparrow$       & 0.886 &\cellcolor{gray!10}\textbf{0.898}      \\
\midrule
\multirow{2}{*}{\textit{HKU}}  & $M$~$\downarrow$       & 0.022 &\cellcolor{gray!10}\textbf{0.021}      \\
& $F_\beta$~$\uparrow$       & 0.927 &\cellcolor{gray!10}\textbf{0.935}      \\
\bottomrule 
\end{tabular}}
\label{table:LDMSOD} 
\end{subtable}
\begin{subtable}[c]{0.356\textwidth}
\centering
\setlength{\abovecaptionskip}{0cm}
\setlength{\belowcaptionskip}{0.05cm}
\caption{Image dehazing}
\resizebox{\columnwidth}{!}{
\setlength{\tabcolsep}{1.1mm}
\begin{tabular}{l|c|cc}
\toprule
Datasets                       & Metrics  & CORUN~\cite{fang2024real} &\cellcolor{gray!10} CORUN+DR-LDM \\ \midrule
\multirow{4}{*}{\textit{RTTS}} & FADE~$\downarrow$     & 0.824 &\cellcolor{gray!10} \textbf{0.785}  \\
& BRISQUE~$\downarrow$  & 11.96 &\cellcolor{gray!10} \textbf{10.63} \\
& MUSIQ-K~$\uparrow$  & 63.82 &\cellcolor{gray!10} \textbf{69.57}  \\
& CLIP-IQA~$\uparrow$ & 0.683 &\cellcolor{gray!10} \textbf{0.752}  \\
\bottomrule
\end{tabular}}
\label{table:LDMhazing} 
\end{subtable}
\end{minipage}\vspace{-6.5mm}
\end{table*}

\noindent \textbf{Effect of OCFormer}. As shown in~\cref{table:Ablation}, removing any of the core components, including DRA, PDR, or DR‑LDM, leads to a performance drop, highlighting the importance of these modules. Also, results in~\cref{fig:VisualAblation} verify that adding DR-LDM helps color correction and detail highlight.

\noindent \textbf{Two-phase training strategy.} Two-phase training is essential for DR-LDM to approximate the oracle prior space established by PI. As shown in~\cref{table:Ablation}, it significantly outperforms one-phase training. Comparing Phase~II strategies, joint fine-tuning of all components surpasses training DR-LDM alone with a frozen backbone (w/o joint), confirming stable parameter transfer and the benefit of end-to-end optimization. \cref{fig:convergence} plots the PSNR curves (averaged over 5 runs with error bars) across 200k total iterations (150k for Phase~I, 50k for Phase~II), with an additional 15k iterations for convergence verification. Phase~II initializes from Phase~I and converges rapidly without oscillation, reaching Phase~I-level performance within only 10k iterations.

\noindent \textbf{Hyperparameter configurations}. As shown in~\cref{table:ParaAnalysis,fig:stagewise}, performance improves with more unfolding stages. To balance accuracy and efficiency, we set $K$ as 3. For the timestep $T$ of DR-LDM, we notice that $T=3$ already yields reliable priors, and a similar trade‑off is observed for the prior length $C_p$.

\vspace{-2mm}
\subsection{Further Analysis and Applications}
\vspace{-1mm}

\noindent \textbf{Importance of combining DUNs with priors.}
We design a baseline in which DR-LDM encodes only the low-quality input $\mathbf{y}$ as the prior to guide UnfoldLDM. As shown in \cref{table:ParaAnalysis} (vector length $C_p^1$), this configuration is highly sensitive to the prior vector length and yields substantially inferior results, as priors derived solely from degraded inputs are corrupted by spatially correlated artifacts (\cref{fig:EffectDR-LDM}). In contrast, when 
DR-LDM receives the MGDA-refined estimates 
$\hat{\mathbf{x}}_k$ and $\tilde{\mathbf{x}}_k$, the stage-wise visualization in \cref{fig:stagewise,fig:VisualizationStage} reveals that the unfolding architecture produces progressively cleaner intermediate estimates across stages, which in turn enable DR-LDM to generate increasingly reliable priors. This bidirectional reinforcement, where the prior guides DUN-based restoration and the DUN refines prior generation, is the key to our framework's effectiveness.

\noindent \textbf{Computational efficiency and lightweight variant.} \cref{table:efficiency} reports a comprehensive efficiency comparison including parameters, FLOPs, GPU memory, training time, and inference time, along with both distortion and perceptual quality metrics. Our base model (UnfoldLDM) achieves +0.39~dB higher PSNR and substantially better perceptual scores (FID~\cite{heusel2017gans} 10.65 vs.\ 13.26, MANIQA~\cite{yang2022maniqa} 0.487 vs.\ 0.455) compared to Reti-Diff, while training 40\% faster (18.59h vs.\ 30.89h) and running at comparable inference speed. When the degradation type is known, we further propose UnfoldLDM-l, a lightweight variant that fixes the MGDA degradation model using Retinex theory and adopts reduced configurations ($K{=}2$, $C_p{=}16$, quarter-channel DR-LDM and OCFormer). UnfoldLDM-l achieves +0.65~dB higher PSNR than CIDNet with 44\% fewer parameters and 72\% fewer FLOPs, while also delivering better perceptual quality (FID 15.03 vs.\ 18.60), confirming the scalability of our framework from full-capacity degradation-invariant models to compact task-specific ones.

\noindent \textbf{User study}. 
We assess the perceptual quality across low-light (\textit{L-v1} and \textit{L-v2}), underwater (\textit{UIEB}), and backlit (\textit{BAID}) enhancement tasks.
Twelve participants with experience in image quality assessment rated each enhanced image on a five‑point scale (1 = worst, 5 = best) based on noise/artifacts, structural preservation, and color fidelity.
Each degraded image and its enhanced version were displayed side‑by‑side in randomized order.
As shown in~\cref{table:UserStudy}, UnfoldLDM receives the highest average scores, demonstrating superior perceptual quality.

\noindent \textbf{Benefits for downstream applications}. We evaluate the impact of our enhanced images on low‑light object detection.
Following \cite{he2023reti}, enhanced results from each method are applied to \textit{ExDark} with YOLO retrained accordingly.
As reported in \cref{table:Detection}, our method achieves the best detection accuracy, confirming that improved restoration quality benefits downstream vision tasks.

\noindent \textbf{Generalization of UnfoldLDM}. We further evaluate the generalizability of UnfoldLDM by integrating the proposed DR‑LDM into various DUNs across multiple tasks.
In this setting, DR‑LDM serves as a coarse‑to‑fine prior generator that provides task‑adaptive latent priors to guide the multi‑stage optimization process.
As reported in~\cref{table:generalization}, consistent performance gains are observed across six representative DUN‑based models on their respective tasks, demonstrating both the versatility and superiority of our framework. 
It is worth note that the improvements are consistent across both low-level tasks (restoration, enhancement, fusion) and the high-level task (salient object detection), suggesting that DR-LDM provides a universal structural prior that benefits diverse vision architectures beyond image reconstruction.

\vspace{-2mm}
\section{Discussion}\vspace{-1mm}
UnfoldLDM integrates the model-based interpretability of DUNs with the generative capability of LDMs, contributing a scalable paradigm to both DUN theory and prior-guided restoration:

From the DUN perspective, UnfoldLDM overcomes two core limitations of the proximal-gradient scheme. First, the MGDA module replaces the fixed degradation operator with a data-driven estimation of both holistic $\mathbf{D}$ and decomposed $(\mathbf{W}, \mathbf{M})$, enabling robust modeling of unknown degradations without sacrificing interpretability. Second, the decoupled proximal design, where DR-LDM extracts a compact, degradation-resistant prior $\mathbf{P}_k^h$ to guide OCFormer, explicitly recovers high-frequency textures suppressed by gradient updates, effectively resolving the over-smoothing bias inherent in existing DUNs.

From the prior-guided restoration perspective, UnfoldLDM introduces iterative prior refinement within multi-stage optimization. As unfolding progresses, DR-LDM receives progressively cleaner estimates $(\hat{\mathbf{x}}_k, \tilde{\mathbf{x}}_k)$, producing increasingly robust priors. Crucially, $\mathbf{P}_k^h$ serves not as a passive regularizer but as an active conditional signal within OCFormer, guiding the recovery of fine-grained textures and maximizing its contribution to structural preservation.

In summary, UnfoldLDM provides a modular template for model-based restoration: MGDA can serve as a plug-and-play upgrade granting existing DUNs blind restoration capability, while the iterative LDM integration offers a new methodology for incorporating generative priors across diverse low-level vision tasks.

\vspace{-2mm}
\section{Limitations and Future Work}\vspace{-1mm}
UnfoldLDM assumes same-resolution input and output in its MGDA gradient steps, requiring external upsampling for tasks like super-resolution. Additionally, the multi-stage design may face memory constraints for extremely high-resolution images despite parameter sharing.
Future directions include: (i)~compact prior extraction via consistency distillation or flow-based models to reduce diffusion overhead; (ii)~extending UnfoldLDM to video restoration by leveraging temporal coherence across unfolding stages.

\vspace{-2mm}
\section{Conclusion}\vspace{-1mm}
\label{sec:conslusion}
In this paper, we propose UnfoldLDM, the first to integrate DUNs with a LDM, for BIR tasks. This addresses two limitations of existing DUNs: degradation-specific dependency and over-smoothing bias. Specifically, we first introduce MGDA for robust degradation estimation. Then, we employ DR-LDM to extract the degradation-invariant prior and use the prior to guide OCFormer for explicit detail restoration. Abundant experiments comprehensively demonstrate the superiority of our UnfoldLDM
in achieving a leading place.

\bibliographystyle{splncs04}
\bibliography{main}
\end{document}